\documentclass[11pt, a4paper, logo, copyright]{googledeepmind}

\pdftrailerid{redacted}

\usepackage{algorithm}
\usepackage[noend]{algpseudocode}
\usepackage{array}
\usepackage{amsmath}
\usepackage{amssymb}
\usepackage{bbm}
\usepackage{bm}
\usepackage{color}
\usepackage{colortbl}
\usepackage{float}
\usepackage{graphicx}        %
\usepackage{subcaption}
\usepackage[dvipsnames]{xcolor}

\usepackage{sidecap}

\usepackage{tikz}

\usepackage{array}
    \newcolumntype{C}{>{$}c<{$}}
    \newcolumntype{L}{>{$}l<{$}}
    \newcolumntype{R}{>{$}r<{$}}

\usepackage[subpreambles=true]{standalone}

\definecolor{cvprblue}{rgb}{0.21,0.49,0.74}
\usepackage[dvipsnames]{xcolor}
\usepackage{adjustbox}
\usepackage{listings}
\usepackage{multirow}
\usepackage{wrapfig}

\def\abf{\mathbf{a}}
\def\hbf{\mathbf{h}}
\def\Vbf{\mathbf{V}}
\newcommand{\sparc}{SPARse Fine-grained Contrastive Alignment}
\newcommand{\sparcshort}{SPARC}

\makeatletter
\renewcommand\bibentry[1]{\nocite{#1}{\frenchspacing\@nameuse{BR@r@#1\@extra@b@citeb}}}
\makeatother

\usepackage{kantlipsum, lipsum}
\usepackage{dsfont}
\usepackage{gdm-colors}

\usepackage[authoryear, sort&compress, round]{natbib}

\graphicspath{{figures/}}

\title{Improving fine-grained understanding in image-text pre-training}

\correspondingauthor{[ioanab,anastasijailic,msbauer,gokererdogan,matko,kaplanis,agritsenko,mjlm,cblundell,razp,mitrovic]@google.com}

\reportnumber{001} %

\author[1]{Ioana Bica}
\author[1]{Anastasija Ilić}
\author[1]{Matthias Bauer}
\author[1]{Goker Erdogan}
\author[1]{Matko Bošnjak}
\author[1]{Christos Kaplanis}
\author[1]{\\Alexey A. Gritsenko}
\author[1]{Matthias Minderer}
\author[1]{Charles Blundell}
\author[1]{Razvan Pașcanu}
\author[1]{Jovana Mitrović}

\affil[1]{Google DeepMind}

\begin{abstract}

We introduce \sparc{} (\sparcshort), a simple method for pretraining more fine-grained multimodal representations from image-text pairs. Given that multiple image patches often correspond to single words, we propose to learn a grouping of image patches for every token in the caption. To achieve this, we use a sparse similarity metric between image patches and language tokens and compute for each token a language-grouped vision embedding as the weighted average of patches. The token and language-grouped vision embeddings are then contrasted through a fine-grained sequence-wise loss that only depends on individual samples and does not require other batch samples as negatives. 
This enables more detailed information to be learned in a computationally inexpensive manner. SPARC combines this fine-grained loss with a contrastive loss between global image and text embeddings to learn representations that simultaneously encode global and local information. We thoroughly evaluate our proposed method and show improved performance over competing approaches both on image-level tasks relying on coarse-grained information,  e.g. classification, as well as region-level tasks relying on fine-grained information, e.g. retrieval, object detection, and segmentation. Moreover, SPARC improves model faithfulness and captioning in foundational vision-language models.

\end{abstract}    

\makeatletter
\renewcommand\paragraph{\@startsection{paragraph}{4}{\z@}%
                                    {1.25ex \@plus1ex \@minus.2ex}%
                                    {-1em}%
                                    {\normalfont\normalsize\bfseries}}
\makeatother

\definecolor{color_comments}{RGB}{14, 113, 2}
\definecolor{color_parentheses}{RGB}{0, 2, 248}
\definecolor{color_strings}{RGB}{145, 1, 16}
\definecolor{color_literals}{RGB}{19, 85, 53}
\definecolor{color_functions}{RGB}{101, 76, 29}
\definecolor{color_return}{RGB}{157, 0, 210}
\definecolor{color_function_args}{RGB}{0, 0, 109}

\newcommand{\colorparentheses}[1]{\bfseries\textcolor{color_parentheses}{#1}}
\newcommand{\colorliterals}[1]{\bfseries\textcolor{color_literals}{#1}}

\lstdefinestyle{mystyle}{
    commentstyle=\color{color_comments},
    keywordstyle=\color{BlueViolet}\bfseries,
    stringstyle=\color{color_strings},
    basicstyle=\tiny\ttfamily,
    breakatwhitespace=false,         
    captionpos=b,                    
    keepspaces=true,                 
    numbers=left,                    
    numbersep=5pt,                  
    showspaces=false,                
    showstringspaces=false,
    showtabs=false,                  
    tabsize=2,
    xleftmargin=20pt,
    frame=tb,
    framexleftmargin=20pt
}

\begin{document}

\maketitle

\section{Introduction}

Contrastive pre-training from large-scale, noisy image-text datasets~\citep{clip, align} has become a widely used paradigm for learning general vision representations useful for a wide range of downstream tasks as well as for learning vision encoders in multimodal foundation models~\citep{flamingo, pali, blip}. By aligning global image and text representations in a shared latent space using similar and dissimilar image-text pairs, these models achieve impressive performance on image-level vision tasks like classification~\citep{clip}, coarse-grained retrieval and visual question answering~\citep{flamingo, pali}. On the other hand, these models have been shown to discard fine-grained visual information~\citep{krojer2022image} and work poorly on downstream tasks involving localization~\citep{zhong2022regionclip, ranasinghe2022perceptual}, counting~\citep{paiss2023teaching} and understanding spatial relationships between objects~\citep{parcalabescu2021valse} or object attributes~\citep{aro}. These shortcomings are further exacerbated when these pretrained models are used in foundation models~\citep{flamingo, pali, blip} or when they are used to initialize models for object detection~\citep{minderer2022simple} or segmentation~\citep{zhou2022extract}.

A recent line of work has started to explore incorporating losses between image patch and text token embeddings~\citep{filip, pacl, gloria, mgca} to learn representations encoding more fine-grained details. Motivated by the idea of aligning patches corresponding to individual objects in the image to tokens corresponding to the words describing these objects, these local losses learn soft correspondences between image patches and text tokens from image-text pairs.
While these models have achieved improved performance on fine-grained retrieval~\citep{filip}, image classification~\citep{filip}, object detection and segmentation~\citep{mgca, pacl}, they are computationally and memory expensive, unstable during training \citep{filip} and/or rely on pretrained models to kickstart learning.

\begin{wrapfigure}{r}{0.6\textwidth}
  \begin{center}
	\includegraphics[width=0.6\columnwidth]{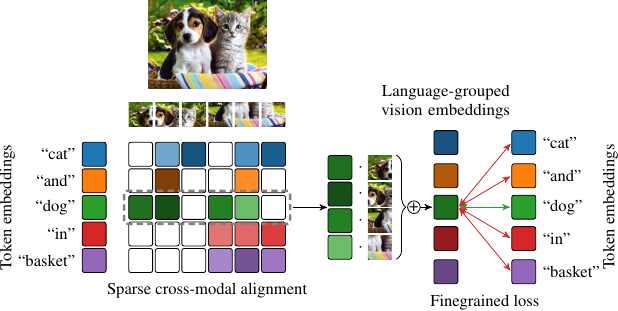}
  \end{center}
  \caption{For every text token, SPARC learns a corresponding language-grouped vision embedding as the alignment-weighted combination of patches that are most similar to that token. We calculate a sparse similarity metric between tokens and patches of individual image-text pairs (left) and use it to compute the resulting alignment weights (middle). We contrast the language-grouped vision embeddings with token embeddings in a fine-grained contrastive sequence-wise loss (right).}
  \label{fig:sparc_intro}
\end{wrapfigure}

In this work, we propose \emph{\sparc{} (\sparcshort)}, a novel objective for multimodal pretraining which learns representations that encode both coarse-grained/global and fine-grained/local information. 
We propose to build \emph{language-grouped vision embeddings} by learning to aggregate (in an unsupervised way) image patches corresponding to individual words in the caption; this is motivated by the observation that usually multiple image patches  correspond to one word in the caption. As a first step, \sparcshort{}  computes the similarity between the patch and token embeddings of an individual image-text pair and enforces sparsity in the resulting similarity matrix. This sparsification enables only the most relevant image patches to be attributed to individual tokens.
Next, as illustrated in Figure \ref{fig:sparc_intro}, for every text token, we compute the corresponding language-grouped vision embedding as the alignment-weighted sum of the patch embeddings, where the alignment weights are computed from the sparsified similarity matrix. 
The resulting language-grouped vision embeddings are contrasted with the token embeddings from the same image-text pair by optimizing for the similarity between individual tokens and their corresponding language-grouped vision embedding and dissimilarity to all other language-grouped vision embeddings.
\sparcshort{} combines the resulting fine-grained/local contrastive loss with a global contrastive loss between image and text embeddings which enables it to simultaneously encode global and local information in the learned representations.

Through its design choices, SPARC addresses several shortcomings of existing methods for learning image representations with more fine-grained information. 
Firstly, several of these methods~\citep{filip, pacl, gloria} learn representations with fine-grained losses that compute similarities between all image patch embeddings and all text token embeddings in a batch. This approach is both computationally and memory intensive and does not scale to large batch sizes (which are needed for obtaining good performance for contrastive methods~\citep{clip, align, sigclip}). On the other hand, SPARC contrasts patch and token embeddings at the level of individual image-text pairs and does not use other examples from the batch to compute the similarity matrix which leads to more favourable computation and memory footprints and more easily scales to large batch sizes.
Secondly, for learning soft correspondences between image patches and text tokens, prior work~\citep{pacl, gloria, mgca} usually relies on building cross-modal weighted representations with weights computed as a softmax over patch and token embedding similarities. The winner-takes-all dynamics of softmax~\citep{Peterson89,elfadel1993softmax} strongly bias learning towards one-to-one mappings between individual text tokens and image patches which often does not correspond to underlying data.  
For example, in an image of a dog, the token embedding for ``dog'' should be matched with \emph{all} patch embeddings that correspond to the dog in the image and not just one/a few. 
Moreover, softmax can be problematic from a gradient flow perspective~\citep{hoffmann2023softmax,shen2023softmax,zhai2023softmax} as it tends to lead to a low entropy distribution, where softmax \emph{saturates} and therefore its Jacobian vanishes~\citep{hoffmann2023softmax}. See Appendix~\ref{app.softmax} for a more detailed explanation. 
On the flip side, SPARC does not use softmax for calculating the alignment weights which allows it to learn a flexible one-to-many matching between individual tokens and the corresponding image patches and to avoid the winner-take-all dynamics of softmax. 
Thirdly, several of these approaches start from contrastively pre-trained vision-language models~\citep{pacl} or from pre-trained language models \citep{gloria, mgca}. %
Moreover, existing fine-grained objectives have been developed in different communities (i.e. medical~\citep{gloria, mgca} vs. general vision~\citep{filip, pacl}) leveraging different types and sizes of datasets, architectures and pretraining setups. This makes it difficult to compare different approaches and assess the benefits of using individual fine-grained objectives.

To summarize, our main contributions are as follows:
\begin{itemize}
    \item  We propose SPARC, a novel method for pre-training multimodal models on large-scale noisy image-text data which learns both coarse-grained and fine-grained information.
    \item  Through an extensive experimental evaluation, we show that SPARC significantly improves performance on both fine-grained and coarse-grained downstream tasks over competing methods. 
    \item  For the first time in the literature, we perform a thorough like-for-like comparison on the benefits of different fine-grained objectives for large-scale pretraining of multimodal models.
\end{itemize}

\section{Sparse Fine-grained Contrastive Alignment} \label{sec.sparc}

Let $\mathcal{B} = \{(\bm{x}_1^v, \bm{x}_ 1^t), (\bm{x}_2^v, \bm{x}_2^t), \dots, (\bm{x}_B^v, \bm{x}^t_B) \}$ be a mini-batch of image-text pairs. Let $f_v(\cdot)$ be the image encoder, $f_t(\cdot)$ the text encoder and  $g_v(\cdot)$ and $g_t(\cdot)$ linear adaptors. %
For an image $\bm{x}^v_i$, we denote the corresponding patches as $(\bm{x}^{v}_{i,1}, \bm{x}^{v}_{i,2}, \dots, \bm{x}^{v}_{i,P})$ and the patch embeddings as $(\bm{v}_{i,1}, \bm{v}_{i, 2}, \dots, \bm{v}_{i,P})$ with $\bm{v}_{i,p}  = g_v(f_v(\bm{x}^{v}_{i, p})) \in  \mathbb{R}^{d}$; $P$ denotes the number of patch embeddings.  We calculate the global vision embedding as $\overline{\bm{v}}_i = g_v(h_v(\text{avg\_pool}( \{f_v(\bm{x}^{v}_{i,p})\}_{p=1}^P )))$ with $h_v$ being a single non-linear layer that facilitates the encoding of different granularities of information.
For the corresponding text $\bm{x}^t_i$, we denote the tokens as $(\bm{x}^{t}_{i,1}, \bm{x}^{t}_{i,2}, \dots, \bm{x}^{t}_{i,L_{i}})$ with $L_i$  the number of tokens for sample $i$. The token embeddings $(\bm{t}_{i, 1}, \bm{t}_{i, 2}, \dots, \bm{t}_{i, L_i})$ are computed as $\bm{t}_{i, l} = g_t(f_t(\bm{x}^t_{i, l}))$ and the global text embedding $\overline{\bm{t}}_i$ is computed by average pooling $\{f_t(\bm{x}^{t}_{i,l})\}_{l=1}^{L_{i}}$ and applying the adaptor $g_t$, i.e. $\overline{\bm{t}}_i = g_t(\text{avg\_pool}( \{f_v(\bm{x}^{t}_{i,l})\}_{l=1}^{L_i})$.

\begin{figure*}[t]
	\centering
    \includegraphics[width=1.0\columnwidth]{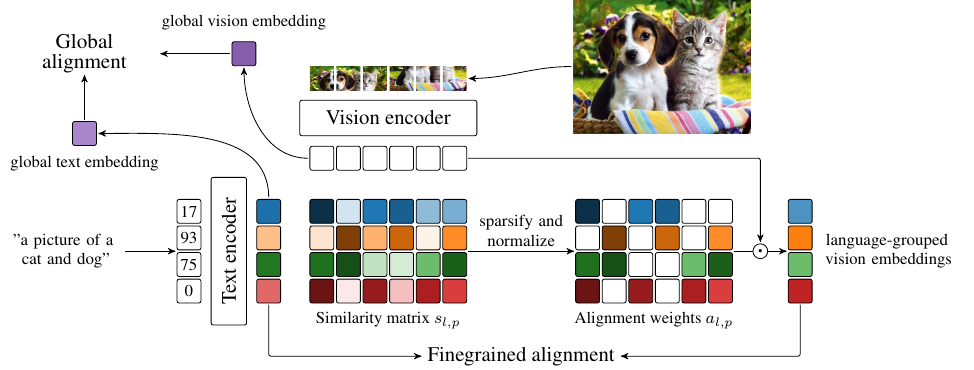}
	\caption{Overall architecture for SPARC. The global alignment loss maximizes the similarity between the global vision and global text embeddings, while minimizing the similarity with the other global embeddings in the batch. To obtain the finegrained alignment, we compute the similarity between the patch embeddings and the token embeddings and then sparsify and normalize the resulting similarity matrix to obtain alignment weights. These alignment weights are then used to group the patch embeddings. The resulting language-grouped vision embeddings are then contrasted to the token emebddings in a sequence-wise finegrained alignment loss.  }
	\label{fig:sparc}
\end{figure*}

\paragraph{Global alignment:} In order to learn global information, SPARC uses the global contrastive loss~\citep{clip, align} which operates at the level of global image ($\overline{\bm{v}}$) and global text embeddings ($\overline{\bm{t}}$). Specifically, we learn image and text embeddings by maximizing the similarity to the corresponding text and image embeddings, while minimizing the similarity to other text and image embeddings in the batch, i.e. we optimize

\begin{equation}
L_g =   - \frac{1}{2B} \sum_{i=1}^{B} \left( \log \frac{  \exp(\phi(\overline{\bm{v}}_i, \overline{\bm{t}}_i) / \tau)}{\sum_{j=1}^B\exp(\phi(\overline{\bm{v}}_i, \overline{\bm{t}}_j) / \tau)} \right. + \left. \log \frac{\exp(\phi( \overline{\bm{t}}_i, \overline{\bm{v}}_i) / \tau)}{\sum_{j=1}^B\exp(\phi(\overline{\bm{t}}_i, \overline{\bm{v}}_j) / \tau)} \right),
\end{equation}
with $\phi(\overline{\bm{v}}_i, \overline{\bm{t}}_j) = \tfrac{\bar{\bm{v}}_i}{\| \bar{\bm{v}}_i \|_2} \cdot  \tfrac{\bar{\bm{t}}_j}{\| \bar{\bm{t}}_j \|_2}$ and $\tau$ as  temperature.

\paragraph{Finegrained alignment:}
Motivated by the observation that usually multiple image patches correspond to one word in the caption, we propose to learn groupings of patches that correspond to individual text tokens.
Specifically, for every token embedding we learn a corresponding \emph{language-grouped vision embedding} as an alignment-weighted combination of patches that encode that token in the visual domain. 
We propose to compute the alignment weights based on the similarity between token and patch embeddings of the corresponding image-text pair. 
To facilitate the grouping of appropriate patch embeddings given a text token we sparsify and min-max normalize the similarity matrix to compute the alignment weights. 
To learn language-grouped vision embeddings, we propose a fine-grained local loss that optimizes for the alignment between individual token embeddings and their corresponding language-grouped vision embeddings within a given image-text pair. 
Specifically, we propose a sequence-wise contrastive loss to optimize this fine-grained alignment within SPARC.
Optimizing this loss (in addition to the global contrastive loss above) biases the learned representation to preserve detailed information about the image (as described by the caption) instead of just the global information sufficient to minimize the global contrastive loss.

For an image-text pair, let $s_{i, lp}$ %
 represent the similarity between text token embedding $\bm{t}_{il}$ and image patch embedding $\bm{v}_{ip}$, i.e. $s_{i,lp} = \bm{t}_{il} \cdot \bm{v}_{ip} $, where $s_{i,lp} \in \mathbb{R}^{L \times R}$ and $\cdot$ is the inner product. Going forward we drop the example index $i$ for simplicity. To obtain alignment weights, for each token $j$, we first normalize $s_{lp}$ to $[0, 1]$ using min-max normalization across columns (i.e. patches):
\begin{equation}
\hat{s}_{lp} = \frac{s_{lp} - \min_{k} s_{l k}}{\max_{k} s_{lk} - \min_{k} s_{lk}}    
\end{equation}
We sparsify the similarity matrix $S = (\hat{s}_{j k})_{1\leq j \leq L, 1 \leq k \leq P}$ to facilitate learning and to encourage each token to be aligned to a few of the patches, i.e.
\begin{equation}
  \tilde{s}_{j k} =
    \begin{cases}
      \hat{s}_{j k} & \text{if $\hat{s}_{j k} \geq \sigma$} \\
      0 & \text{otherwise} \\
    \end{cases}   
\end{equation}
with $P$ the number of patch embeddings of an image and $\sigma$ the sparsity threshold. 
We compute alignment weights as
\begin{equation}
    a_{j k} = \frac{\tilde{s}_{j k}}{\sum_{r=1}^{R} \tilde{s}_{j r}}
\end{equation}
where $a_{jk}$ represents the weight of patch $k$ for computing the language-grouped vision embedding corresponding to token $j$. 
Note that this approach enables a flexible mapping between a token and arbitrarily many patch embeddings that encode that token in the visual domain, e.g. all of the image patches corresponding to ``dog'' can be matched to the token encoding ``dog''.
For every token $t_{l}$ we compute the corresponding language-grouped vision embedding $\bm{c}_l$ as 
\begin{equation}
    \bm{c}_l = \sum_{r=1}^{R} a_{lr} \bm{v}_r
\end{equation}
as the alignment-weighted combination of patch embeddings with $R$ the number of patches with non-zero alignment weight. 

To learn fine-grained information we propose to optimize the alignment between token embeddings and their corresponding language-grouped vision embeddings. Specifically we propose a fine-grained contrastive loss that operates over sequences of tokens and patches at the level of each image-text pair and does not require negatives from other image-text pairs. 
This considerably reduced computation and memory costs over previous methods~\citep{filip, gloria} that require samples from the whole batch in order to compute their fine-grained losses.
SPARC optimizes the following fine-grained alignment contrastive loss

\begin{footnotesize}
\begin{equation}
L_f =     - \frac{1}{2B} \sum_{i=1}^{B} \left[ \frac{1}{L_i} \sum_{j=1}^{L_i} \left( \log \frac{\exp(\phi(\bm{c}_{ij}, \bm{t}_{i j}) / \tau)}{\sum_{k=1}^{L_i} \exp(\phi(\bm{c}_{ij}, \bm{t}_{ik}) / \tau)} \right. \right. + \left. \left. \log \frac{\exp(\phi(\bm{t}_{ij}, \bm{c}_{ij}) / \tau)}{\sum_{k=1}^{L_i}\exp(\phi(\bm{t}_{ij}, \bm{c}_{ik}) / \tau)} \right) \right], %
\end{equation}
\end{footnotesize}

which tries to maximize the similarity of every token embedding with its corresponding language-grouped vision embedding and minimize the similarity to other language-grouped vision embeddings in the sequence and vice versa.

\paragraph{Overall objective: } The overall SPARC objective is a weighted sum of the global contrastive loss and the finegrained alignment constrastive loss: 
\begin{equation}
    L_{\text{SPARC}} = \lambda_g L_g + \lambda_f L_f
\end{equation}
where $\lambda_g$ and $\lambda_f$ are hyperparameters.  We provide the pseudo-code for SPARC in Appendix~\ref{app.pseudocode}.

\paragraph{Sparsity threshold.} We choose the sparsity threshold $\sigma$ to be equal to $1/P$ with $P$ the number of image patches. This choice is motivated by the consideration that every text token should attend to at least to one image patch. Since we use the min-max normalization the smallest similarity of $1/P$ is achieved when all patches are equally similar as the number of patches is constant.
Note that this threshold naturally allows for the number of patches corresponding to one token to considerably vary between tokens within an image as well as across images; this enables the same class of objects (e.g. ``dogs'') to be appropriately represented irrespective of the difference in sizes, scales and shapes across different instances within and across images.
Note also that the threshold also allows for the decoupling of similarities of individual patches to different tokens as it allows for different number of zero entries in different rows of the similarity matrix; thus, whether and how much a patch is similar to a token, has no bearing to how similar it is to a different token which is useful e.g. in situations when we have more detailed captions (e.g. ``large brown dog'')  and/or when a single word is represented by multiple tokens.

\section{Related work} \label{sec:related_work}

\paragraph{Contrastive image-text pre-training} CLIP~\citep{clip} and ALIGN~\citep{align} popularized learning general visual representations by leveraging textual supervision from noisy large-scale data scrapped from the internet. These methods learn representations  through a contrastive objective that maximises the similarity between the representation of the whole image and the representation of the full text of matched image-text pairs and minimizes the similarity between the remaining image-text pairs within the batch. %
However, learning visual representations through matching the global image and text embeddings can result in a coarse visual representation that discards many fine-grained details (i.e all details that are not needed for differentiating the matching of global text embedding from the other text embeddings in the batch). To address this problem, FILIP~\citep{filip} proposes a \emph{cross-modal late interaction mechanism}, which optimizes the token-wise maximum similarity between image and text tokens through  a contrastive objective. While this approach achieves a finer-grained alignment between image patches and words in the text, computing the token-wise similarity between all image patches and text tokens in the batch becomes memory inefficient for large batch sizes so they use several tricks during pre-training to address this issue. A related approach PACL~\citep{pacl} starts from CLIP-pretrained vision and text encoders and trains on top of the frozen  representations an adapter to obtain better fine-grained understanding. The adapter is a two-layer MLP with a residual connection and is trained through a contrastive objective that compares the global text embedding and a weighted global image embedding with the weights calculated using the cosine similarity between individual image patches and the global text embedding.

In a parallel stream of work, several methods have been proposed in the medical literature to learn visual representation using medical images - radiology report pairs from small scale datasets (consisting of up to ~200k data points)~\citep{gloria, mgca, Dawidowicz2023LIMITRLL}. GLoRIA~\citep{gloria}   builds localized visual representations by contrasting attention-weighted patch embeddings with the text tokens, where the attention weights are computed through softmax on the similarity matrix between the patch and token embeddings. Similarly to FILIP, the local objective in GLoRIA requires computing the similarity between all patch and token embeddings within the batch which is computationally intensive and does not scale to large batch sizes. Alternatively, MGCA~\citep{mgca} considers a token-wise fine-grained loss that employs a bidirectional multi-head attention strategy to learn the matching between image patch and token embedding. While this is more efficient to compute, learning these matchings through a bidirectional multi-head cross-attention strategy adds more parameters to the dual encoders, involves tuning several additional hyperparameters and suffers from the same problems with using softmax for computing the attention weights. MGCA also uses a domain-specific disease-level alignment loss that enforce a cluster assignment consistency to leverage inter-subject semantic correspondences. More recent methods \citep{Dawidowicz2023LIMITRLL} consider incorporating into the pre-training objective not only fine-grained losses similar to the ones used in GLoRIA and MGCA, but also domain-specific features and image views.  Note that these methods from the medical literature start from a text encoder pre-trained with medical texts~\citep{alsentzer2019publicly}, while we consider the case of pre-training the image and text encoders jointly from scratch.

\paragraph{Fine-grained understanding in vision-language models} Alternative approaches for improving the fine-grained capabilities of vision-language models require pre-trained modules, specialised networks and human annotations. One line of work, proposes matching image regions to textual descriptions through contrastive losses, where the image regions - text description pairs are obtained from human annotations~\citep{li2022grounded} or by using region proposal networks~\citep{ren2015faster} and various text matching approaches~\citep{zhong2022regionclip, varma2023villa}. A separate line of work adds a cross-modal encoder (with significant extra parameters) on top of the dual image-text encoder and uses captioning~\citep{coca, blip}, masked language modelling~\citep{albef, yang2022vision}, image-text matching~\citep{xvlm, albef, yang2022vision} and bounding box prediction losses~\citep{xvlm} (with bounding boxes  obtained from human-annotations~\citep{krishna2017visual, kuznetsova2020open, shao2019objects365}). For more related works see Appendix~\ref{app.related_works}.

\section{Experiments} \label{sec:experiments}

While there has been significant interest in learning fine-grained representations, the breadth of training setups used in the literature have made it difficult to compare different fine-grained objectives. Specifically the use of custom datasets~\citep{filip} and pretrained language and/or vision models~\citep{gloria, mgca, pacl} have made it difficult to discern the benefit of individual fine-grained losses on learning more detailed representations. %
In this work we want to enable a like-for-like comparison and understand the impact of SPARC and competing fine-grained losses on downstream performance. For this purpose, we reimplement all competing baselines: CLIP~\citep{clip}, FILIP~\citep{filip}, PACL~\citep{pacl}, MGCA~\citep{mgca} and GLoRIA~\citep{gloria}, and use the same pretraining datasets, architecture and number of training steps when training with the different objectives; we pretrain randomly initialized networks. 
We thoroughly evaluate the learned representations across a broad range of tasks and datasets, ranging from coarse-grained image-level tasks like classification and retrieval to fine-grained tasks like object detection and semantic segmentation. Unlike some competing methods that improve fine-grained understanding at the cost of decreasing coarse-grained task performance, SPARC simultaneously boosts performance over both coarse- and fine-grained tasks across a number of different benchmarks.

\subsection{Experimental setup}

\textbf{Model architectures}
Following the literature, we use Vision Transformers (ViTs)~\citep{vit} as image encoders and Transformers~\citep{transformer} as text encoders. 
We experiment with ViT-B/32, ViT-B/16 and ViT-L/14 and pair them with corresponding language models. See details in Appendix~\ref{app.experiments}. 

\paragraph{Datasets}
We train using large-scale datasets ALIGN~\citep{align}, JFT~\citep{sun2017revisiting, zhai2022scaling} and LTIP (Long Text \& Image Pairs)~\citep{flamingo}. ALIGN has 1.8 billion images paired with noisy alt-text, JFT has of 4 billion images semi-automatically annotated with a class-hierarchy of 30k labels, %
while LTIP  has 312 million higher-quality images - text pairs with richer image captions. See Appendix \ref{app.experiments} for more details.

\paragraph{Pre-training details} We resize images to the $224\times224$ resolution and tokenize the text with a 32k vocabulary sentencepiece tokenizer~\citep{kudo2018sentencepiece} while keeping a maximum number of 55 tokens for each caption. 
We train all models using the AdamW~\citep{loshchilov2017decoupled} optimizer, a cosine learning rate schedule with linear warm-up and weight decay regularization. We use a batch size of 16348 and we pre-train the ViT-B models for 200k steps ($\approx$ 3.2 billion data points) and the ViT-L models for 250k steps ($\approx 4.1$ billion data points). See Appendix~\ref{app.experiments} for more hyperparameter details.

\begin{table*}[h]
\begin{center}
\begin{tabular}{llCCCCCCCCCCCC}  %
\toprule
& Objective &  \text{IN} & \text{IN-V2 Th} & \text{IN-V2 MF} & \text{IN-V2 TI} & \text{IN-R} & \text{IN-C} & \text{IN-A}  & \text{IN-Sketch}  \\
\midrule
\midrule
\parbox[t]{2mm}{\multirow{6}{*}{\rotatebox[origin=c]{90}{ViT-B/32}}}  & CLIP & 66.7 &	66.2 &	58.9 &	71.5 &	63.2 & 42.6 &	15.1 &	51.7  \\   %
& FILIP  & 52.7 &	50.7	& 44.0	& 55.8 & 47.1 & 28.7 & 8.4 & 38.2  \\  %
& PACL & 58.9 & 56.9 & 50.0 & 62.6 & 54.0 & 34.9 & 9.3 & 44.1  \\  %
& GloRIA & 62.8 &	61.5 &	54.3 & 66.7 &	56.7 &	38.4 & 11.2 &	47.5   \\  %
& MGCA & 66.0 &	64.5 &	56.4 &	69.5 &	62.0 & 41.1 &	14.7 &	51.7   \\  %
& SPARC (ours) & {\bf 68.1} &	{\bf 67.0} & {\bf 59.7} &	{\bf 72.0} & {\bf 64.9} & {\bf 44.5} &	{\bf 16.7}	& {\bf 53.2}  \\   %
\midrule
\midrule
\parbox[t]{2mm}{\multirow{6}{*}{\rotatebox[origin=c]{90}{ViT-B/16}}}  & CLIP & 71.6 &	70.9 &	63.7 &	74.8 & 71.1 &  {\bf 48.5}	& 32.2 &	56.8 \\  %
& FILIP  & 56.6 &	55.6	& 48.9	& 59.7 &  54.0  &   33.2   &   14.4   &  43.1  \\  %
& PACL & 61.1 &	59.6 &	52.6 &	64.8 &	56.3 &	36.1 & 12.8 &	45.2 \\
& GloRIA &  67.4 &	66.9 &	59.8 &	71.7 &	66.6 & 43.8 &	24.6 &	54.2  \\  %
& MGCA & 69.6 &	69.3 &	62.2 &	73.6 &	68.8 &	46.1 & 29.0 &	55.0  \\  %
& SPARC (ours) &  {\bf 72.6}  &  {\bf 71.1}  &  {\bf 64.4}  &  {\bf 75.0}  &	{\bf 72.0} &	{\bf 48.5} & {\bf 33.8} &	{\bf 57.3} \\  %
\midrule
\midrule
\parbox[t]{2mm}{\multirow{3}{*}{\rotatebox[origin=c]{90}{ViT-L/4}}} & CLIP  & 77.3 &	75.9 &	69.5 &	79.1 &	78.8 & 59.6 & 	{\bf 52.5} &	64.5 \\
& MGCA & 75.6 &	73.9 &	68.0 &	77.9 &	77.2 & 56.0 &	45.0 &	63.1 \\
& SPARC (ours) & {\bf 78.2}	& {\bf 76.9}	& {\bf 70.6}	& {\bf 80.0} &	{\bf79.3} &	{\bf 59.7} & 51.9 &	{\bf 65.4} \\
\bottomrule
\end{tabular}
\end{center}
\caption{Top-1 accuracy (in \%) of zero-shot classification on ImageNet (IN) and its variants ImageNet-V2 Threshold (IN-V2 Th), ImageNet-V2 Matched Frequency (In-V2 MF), ImageNet-V2 Top Images (IN-V2 TI), ImageNet-R (IN-R), ImageNet-C (IN-C), ImageNet-Sketch (IN-Sketch). }
\label{tab:classification}
\end{table*}

\begin{table*}[h]
\begin{center}
\begin{tabular}{llCCCCCCCCCCCC}  %
\toprule
& Objective &  \text{IN} & \text{IN-V2 Th} & \text{IN-V2 MF} & \text{IN-V2 TI} & \text{IN-R} & \text{IN-C} & \text{IN-A}  & \text{IN-Sketch}  \\
\midrule
\midrule
\parbox[t]{2mm}{\multirow{6}{*}{\rotatebox[origin=c]{90}{ViT-B/32}}}  & CLIP & 69.0 &	68.8 &	60.4 &	73.4 &	62.4 &   44.6 &	15.8 &	52.4    \\   %
& FILIP  &  56.8 &	54.8 &	48.4 &	60.0 &	44.6 & 30.8 &	7.8 &	39.6 \\  %
& PACL &  61.2 &	59.5 &	51.9 &	65.2 &	52.9 & 36.4 &	9.3 &	45.2 \\  %
& GloRIA &  65.9 &	64.8 &	57.0 &	69.6 &	57.4 & 40.7	& 11.7 &	48.7  \\  %
& MGCA &  68.6	& 67.4	& 59.2	& 72.6 &	61.0 & 43.5 &	14.1 &	50.9 \\  %
& SPARC (ours) & {\bf 70.4} &	{\bf 69.6} &	{\bf 62.1} &	{\bf 74.5} &	{\bf 63.2} &	 {\bf 46.5} & {\bf 17.3} &	{\bf 52.7} \\   %
\midrule
\midrule
\parbox[t]{2mm}{\multirow{6}{*}{\rotatebox[origin=c]{90}{ViT-B/16}}}  & CLIP & 73.9 & 	73.6 &	66.1 &	77.1	& 68.8	& 50.4 &	32.5 &	57.3  \\  %
& FILIP  &   61.4	& 61.0 &	53.8 &	65.6 &	53.2 & 35.9 &		14.2 &	45.1 \\ %
& PACL   & 63.3	& 61.7	& 54.4	& 66.8 &	54.1 & 37.3 &	12.9 &	45.4  \\  %
& GloRIA &  70.4 &	70.0 &	62.8 &	74.7 &	65.7 & 46.4 &		25.0	& 54.8  \\  %
& MGCA &   72.7 &	72.7 &	65.3 &	76.3 &	67.6 & 48.4 &		29.8 &	55.5 \\  %
& SPARC (ours) & {\bf 74.7} &{\bf 74.0} &	{\bf 67.1} &	{\bf 77.8} &	{\bf 71.1} & {\bf 51.31} &	{\bf 34.2} &	{\bf 57.9}  \\  %
\midrule
\midrule
\parbox[t]{2mm}{\multirow{3}{*}{\rotatebox[origin=c]{90}{ViT-L/4}}} & CLIP  & 79.2 &	78.5 &	71.8 &	81.6 &	78.5 & {\bf 61.3} &	51.5 &	65.1 \\
& MGCA & 78.0 &	77.4 &	70.5 &	80.6 &	75.2 & 57.9	& 45.5 & 	63.1 \\
& SPARC (ours) & {\bf 79.7} &	{\bf 78.9} &	{\bf 72.6} &	{\bf 81.9} &	{\bf 79.8} & {\bf 61.3}	& {\bf 53.4}	& {\bf 65.9}  \\
\bottomrule
\end{tabular}
\end{center}
\caption{Top-1 accuracy (in \%) of zero-shot classification using prompt ensembling on ImageNet (IN) and its variants ImageNet-V2 Threshold (IN-V2 Th), ImageNet-V2 Matched Frequency (In-V2 MF), ImageNet-V2 Top Images (IN-V2 TI), ImageNet-R (IN-R), ImageNet-C (IN-C), ImageNet-Sketch (IN-Sketch). }
\label{tab:classification_prompt_ensambling}
\end{table*}

\subsection{Zero-shot image classification}
We first evaluate SPARC on the coarse-grained task of zero-shot image classification.
Specifically we test zero-shot classification on ImageNet~\citep{imagenet} and a number of datasets testing for specific capabilities like robustness to perturbations and various distribution shifts; we choose ImageNetV2~\citep{recht2019imagenet}, ImageNet-R~\citep{hendrycks2021many}, ImageNet-C~\citep{hendrycks2019benchmarking}, ImageNet-A~\citep{hendrycks2019natural} and ImageNet-Sketch~\citep{wang2019learning} for this purpose.
We follow a similar protocol to \citep{clip} for the evaluation, and compute results for both one prompt per example (i.e. the class label) in Table~\ref{tab:classification} and when using prompt ensembling in Table~\ref{tab:classification_prompt_ensambling}.  For more details on the evaluation protocol please see Appendix \ref{app.experiments}.
From both Table~\ref{tab:classification} and Table~\ref{tab:classification_prompt_ensambling} we see that SPARC outperforms or matches competing methods in all settings and across different ViT architectures. Specifically, SPARC shows very effective information encoding from larger patches as exhibited by the significant improvements over baselines for ViT B/32, especially on ImageNet-R, -C, -A and -Sketch showcasing the robustness to perturbations and adversarial examples. Moreover, we notice that while prompt ensembling improves performance of all methods on zero-shot image classification (which is in line with the literature) the performance gain from SPARC are still preserved in this evaluation setting. 

Note that PACL~\citep{pacl}, GLoRIA~\citep{gloria} and MGCA~\citep{mgca} were developed with the use of pretrained language and/or vision encoders in mind, whereas here they are tested in a pretraining from scratch setting. 
From Table ~\ref{tab:classification} and Table ~\ref{tab:classification_prompt_ensambling}, we see that in the pretraining setting PACL and GLoRIA underperform CLIP, whereas MGCA shows more competitive performance to CLIP. 
On the other hand, FILIP~\citep{filip}, which was developed as a fine-grained objective for pretraining from scratch, has proven highly unstable to train across a wide range of learning rates and weight decay parameters which lead to decreased performance. This training difficulty has also been noted in the original paper~\citep{filip} (cf. in the Appendix A.3. \emph{"...training is extremely unstable and the Nan loss easily happens."}). In addition to that FILIP uses a number of additional tricks not present in a standard pretraining setup like image augmentations, backtranslation of captions and custom prompt ensembling.

\begin{table*}[h]
\begin{center}
\begin{adjustbox}{max width=1.9\columnwidth}
\setlength\tabcolsep{3pt}
\begin{tabular}{llCCCCCCCCCCCCCCCC}  %
\toprule
 & &  \multicolumn{6}{c}{Flickr30k} & \multicolumn{6}{c}{MSCOCO}  \\
 &  &  \multicolumn{3}{c}{image-to-text} & \multicolumn{3}{c}{text-to-image} & \multicolumn{3}{c} {image-to-text} & \multicolumn{3}{c} {text-to-image} \\
 &  Objective  & \text{R@1} & \text{R@5} & \text{R@10} & \text{R@1} & \text{R@5} & \text{R@10} & \text{R@1} & \text{R@5} & \text{R@10} & \text{R@1} & \text{R@5} & \text{R@10} \\
\midrule
\midrule
\parbox[t]{2mm}{\multirow{6}{*}{\rotatebox[origin=c]{90}{ViT-B/32}}}  & CLIP &	79.2 &	95.1 &	97.2 &	66.5 &	88.0 & {\bf 93.1} &  53.5 &	78.2 &	86.7 &	38.4 &	64.8 &	74.9   \\
& PACL & 65.5 &	86.8	& 92.2	& 49.8 &	76.5 &	84.7	 & 37.6 &	65.1  & 75.7 &	26.5  &	50.6 &	61.8	  \\   %
& GLoRIA &	74.6 &	92.1 &	96.2 &	61.5 &	85.3 &	90.7 & 46.9 &	73.0 &	82.7 &	34.5 &	61.0 &	71.7  \\
& MGCA &	81.5 &	93.9 &	96.8 &	64.4 &	86.5 &	92.0 & 54.5 &	78.6 &	86.8 &	37.7 &	63.7 &	74.0  \\
& FILIP  & 62.6 & 86.9 &	92.9 &	50.5 &	77.7 &	84.9 &  35.6 &	61.0 &	73.1	& 26.2 & 51.0 & 62.4	\\    %
& SPARC (ours)  &	{\bf 82.5}	& {\bf 96.2} &	{\bf 97.6} &	{\bf 67.7} &	{\bf 88.2}  &	93.0 & {\bf 55.0} &	{\bf 79.1} &	{\bf 87.3} &	{\bf 39.7} &	{\bf 65.9} &	{\bf 75.7} \\
\midrule
\midrule
\parbox[t]{2mm}{\multirow{6}{*}{\rotatebox[origin=c]{90}{ViT-B/16}}}  & CLIP &	84.0 &	96.1 &	98.2 &	71.6 &	90.3 & 94.1  & 56.2 &	80.6 &	88.2 &	42.4 &	{\bf 68.6} &	78.3  \\
& PACL   &	69.6 &	89.7 &	94.2 &	54.9 &	80.7 & 87.3	& 41.8 &	67.8 &	77.6 &	29.1 &	54.3 &	65.5  \\  %
& GLoRIA & 78.0 &	95.5 &	98.0 &	68.4 &	88.9 &	93.2 & 49.7	 & 75.4 &	84.6	& 38.9 &	65.1 &	75.2	 \\
& MGCA &	82.2 &	96.1 &	98.1 &	67.7 &	88.5 &	93.2 & {\bf 57.6} &	80.5 &	87.8 &	39.8 &	65.7 &	75.3  \\
& FILIP  & 69.0	& 89.8	& 94.0	& 55.8	& 81.5 & 	87.9 & 40.2 &	66.0 &	76.3 & 	29.5 & 	55.3	& 66.3		 \\
& SPARC (ours)  &	{\bf 84.4} &	{\bf 97.6} &	{\bf 98.7} &	{\bf 72.0} &	{\bf 91.2} &	{\bf 94.9} & {\bf 57.6} &	{\bf 81.2} &	{\bf 88.5} &	{\bf 43.0} &	{\bf 68.6} &	{\bf 78.5} \\
\midrule
\midrule
\parbox[t]{2mm}{\multirow{3}{*}{\rotatebox[origin=c]{90}{ViT-L/14}}}
& CLIP  &   84.7	& 96.9	& 98.4	& 73.7	& \textbf{91.8}	& \textbf{95.4} & 58.6 &	82.6 &	89.1 &	44.8 &	70.5	& 79.5	  \\
& MGCA  &	85.9 &	96.9 &	98.1 &	73.2 &	91.6	& 95.3 &  \textbf{59.7} &	\textbf{83.2} &	\textbf{89.7} &	44.3 &	69.6 &	78.8 	\\
& SPARC (ours) & \textbf{86.9}	& \textbf{97.3}	& \textbf{98.6}	& \textbf{74.4}	& 91.7	& \textbf{95.4} & 58.9 &	82.9 &	\textbf{89.7 } &	\textbf{45.6}	& \textbf{71.1}	& \textbf{80.1}	 \\
\bottomrule
\end{tabular}
\end{adjustbox}
\end{center}
\caption{Results on zero-shot image-to-text and text-to-image retrieval on MSCOCO and Flickr30k datasets. R@i denotes Recall at i.}
\label{tab:retrieval}
\end{table*}

\subsection{Image-Text retrieval}
Next we evaluate SPARC on zero-shot cross-modal retrieval tasks, i.e image-to-text and text-to-image retrieval, on Flickr30k~\citep{plummer2015flickr30k} and MSCOCO~\citep{lin2014microsoft}. 
From Table \ref{tab:retrieval}, we see that SPARC outperforms all competing baselines across all metrics. 
While using fine-grained losses PACL and GLoRIA significantly underperforms the global contrastive objective CLIP, MGCA shows competitive performance to CLIP in the pretraining setting.
Unfortunately, FILIP~\citep{filip} again underperforms CLIP across all metrics. In an attempt to stabilize FILIP we combined it with CLIP and observed an improvement on image-to-text Flikr30k on ViT B/32 while being competitive on other benchmarks to CLIP. We provide these results in Appendix \ref{app.experiments}.

\subsection{Evaluating faithfulness}
We further examine fine-grained performance of SPARC through \emph{faithfulness}---how consistent the model's highest scoring caption is with the ground truth caption(s)~\citep{ji2023survey}. This is different from top-1 retrieval (R@1) which measures exact match retrieval and does not evaluate the ability of the models to faithfully describe the elements in the image. 
Faithfulness has been used in the LLM literature to assess the propensity of the model to hallucinate~\citep{adlakha2023evaluating,razumovskaia2023textitdial} as models with higher faithfulness more accurately capture the details of the ground truth while not inserting additional information (possible hallucinations). 
The lexical overlap metric of $\mathcal{K}$-Precision measuring the proportion of tokens in the top chosen caption that appear in the ground truth tokens has been shown to correlate well with human judgement ~\citep{adlakha2023evaluating}.
In Table~\ref{table.k_precision} we report the $\mathcal{K}$-Precision on the MSCOCO  for all tokens ($\mathcal{K}$-P), as well as $\mathcal{K}$-Precision restricted to nouns and adjectives only ($\mathcal{K}$-P\textsubscript{na}), as these better encode the objects observed in the image. 
We evaluate all methods on two architectures and see that SPARC reduced hallucinations of objects (higher $\mathcal{K}$-P\textsubscript{na}) while also showing competitive performance to related methods when taking all tokens into account (as measured by $\mathcal{K}$-P).

\begin{table}[H]
 \begin{center}
 \begin{adjustbox}{max width=\columnwidth}
\begin{tabular}{l C C C C}
\toprule
& \multicolumn{2}{c}{ViT-B/32} & \multicolumn{2}{c}{ViT-B/16} \\
Method &  \text{$\mathcal{K}$-P}\textsubscript{na} & \text{$\mathcal{K}$-P} & \text{$\mathcal{K}$-P}\textsubscript{na} &  \text{$\mathcal{K}$-P} \\
\midrule
\midrule
CLIP  & 76.03 &	77.82 & 77.56 &	 78.99  \\
FILIP & 63.3 & 66.83 & 66.05 &	70.09\\
PACL &  3.36 &	26.26  & 4.09 &	27.31 \\
GLoRIA & 71.63 & 73.54 & 73.85 &	75.3  \\
MGCA & 75.79 &	77.98  & 77.66 & \textbf{80.03} \\
SPARC (ours) &  \textbf{76.46} &	\textbf{78.44} & \textbf{78.72} &	79.77 \\
\bottomrule
\end{tabular}
\end{adjustbox}
 \end{center}
\caption{All-token $\mathcal{K}$-Precision ($\mathcal{K}$-P) and the $\mathcal{K}$-Precision restricted to nouns and adjectives ($\mathcal{K}$-P\textsubscript{na}) (in \%) on MSCOCO.}
 \label{table.k_precision}
 \end{table}

\subsection{Fine-grained localization}

We further examine SPARC by evaluating it on fine-grained tasks requiring precise localization such as open-vocabulary object detection and zero-shot semantic segmentation. For these evaluations, we use the ViT-B/16 architecture.

\paragraph{Open-vocabulary object detection.}
To first evaluate whether the improved fine-grained understanding learned with SPARC translates to tasks requiring fine-grained localization, we use SPARC as a backbone for object detection. Specifically, we used the OWL-ViT open-vocabulary object detector~\citep{minderer2022simple} with a ViT-B/16 backbone. After SPARC pre-training, detection heads are added to the backbone and fine-tuned on Objects365~\citep{shao2019objects365} and Visual Genome~\citep{krishna2017visual} datasets following the approach in \citet{minderer2022simple}. 
We evaluate the resulting model on the large-vocabulary dataset LVIS~\citep{gupta2019lvis} which is well-suited for testing the transfer of knowledge from image-level pretraining.
LVIS contains 1203 categories of objects, of which 307 ``rare'' categories are excluded from the training data to measure zero-shot transfer from pretraining. Moreover, we also evaluate detection on the 80 MSCOCO classes.
We run detection training three times and report mean and standard deviation in Table \ref{tab:detection}. SPARC improves over CLIP $+0.9\%$ on LVIS and MSCOCO as measured by mean average precision and $+3.1\%$ on LVIS ``rare'' classes. Since LVIS ``rare'' classes are never seen during detection training data, the model has to rely on information transfer from the pretrained representations for these classes. 
The large improvement of SPARC over the baseline on LVIS $\text{AP}_\text{rare}$  suggests that SPARC has learned more informative fine-grained representations.

\begin{table}[h]
    \centering
 \begin{adjustbox}{max width=\columnwidth}
    \begin{tabular}{lCCC}
         \toprule
           &  \multicolumn{2}{c}{LVIS}& \text{MSCOCO} \\
         Method &  \text{AP}\textsubscript{all} & \text{AP}\textsubscript{rare} & \text{AP}\textsubscript{all}\\
         \midrule
         \midrule
         CLIP & 26.9 \pm 0.12 & 22.0 \pm 0.79 & 38.5 \pm 0.19 \\
         SPARC (ours) & {\bf 27.9\pm 0.11}  &  {\bf 25.1 \pm 0.95} & {\bf 39.4 \pm 0.13} \\
         \bottomrule
    \end{tabular}
 \end{adjustbox}
    \caption{Mean Average precision (as mean $\pm$ standard deviation)  on all and rare classes on LVIS and on all classes in MSCOCO.}
    \label{tab:detection}
\end{table}

\begin{wraptable}{r}{7.5cm}
    \centering
 \begin{adjustbox}{max width=0.5\columnwidth}
    \begin{tabular}{lCC}
         \toprule
        Method & \text{Pascal VOC} & \text{Pascal Context} \\
         \midrule
        \text{CLIP} & 23.02 & 20.45 \\
         \text{FILIP} & 19.32 & 9.31 \\
         \text{PACL} &  1.23 & 1.61 \\
         \text{GLoRIA} & 22.64 & 15.26 \\
         \text{MGCA} & 21.91  & 11.50 \\
         \text{SPARC (ours)} & \mathbf{27.36} & \mathbf{21.65} \\
    \bottomrule
    \end{tabular}
\end{adjustbox}
    \caption{Semantic Segmentation: mIoU of predicted and ground-truth segmentation on Pascal VOC and PASCAL Context datasets.}
    \label{tab:segmentation}
\end{wraptable}

\begin{figure}[t]
    	\centering
    	\subfloat{{\includegraphics[height=2.75cm]{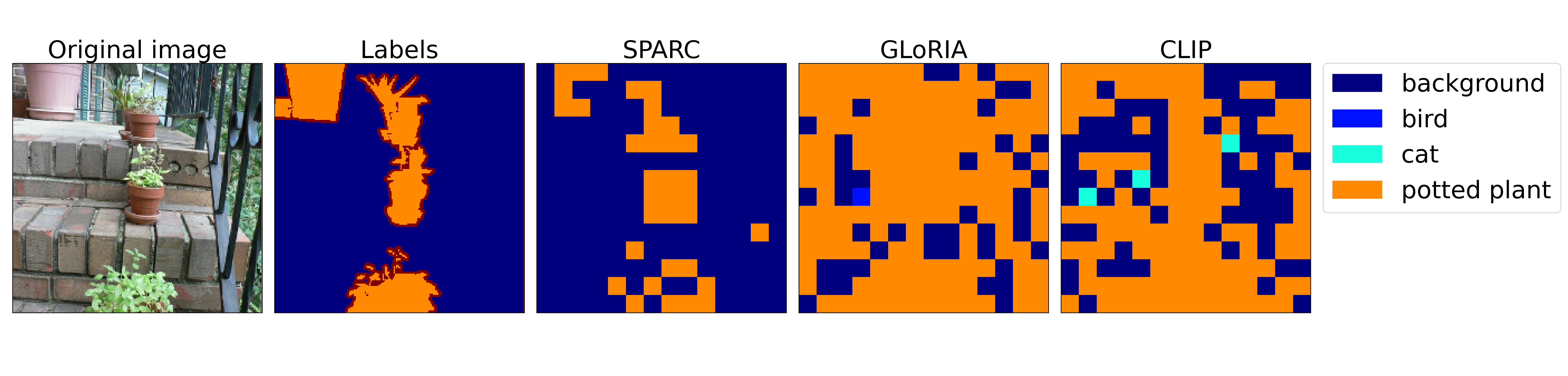} }}%
    	\vspace{-3mm}
    	\quad
    	\subfloat{{\includegraphics[height=2.75cm]{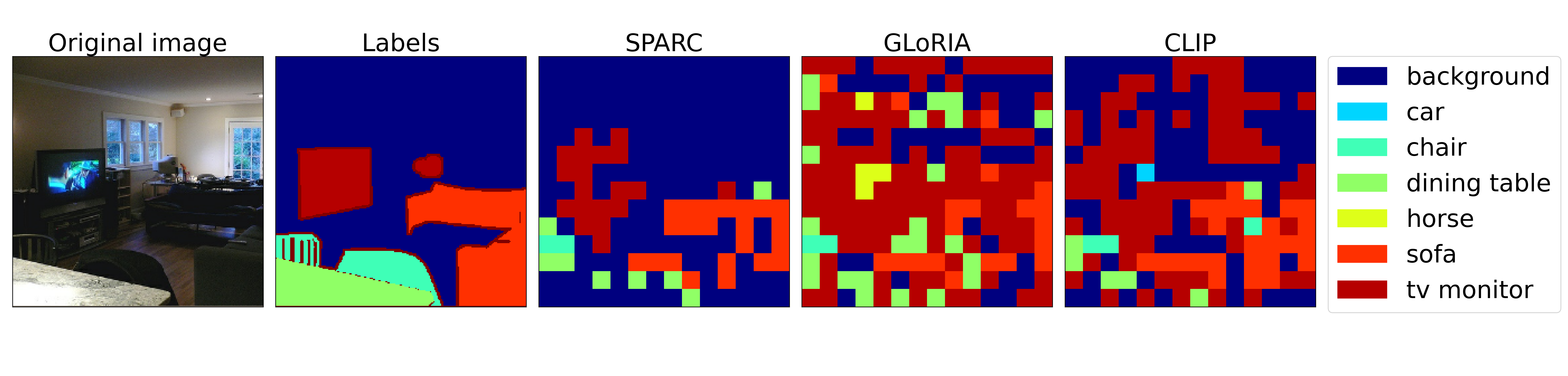} }} 
    	\vspace{-3mm}
    	\quad
    	\subfloat{{\includegraphics[height=2.75cm]{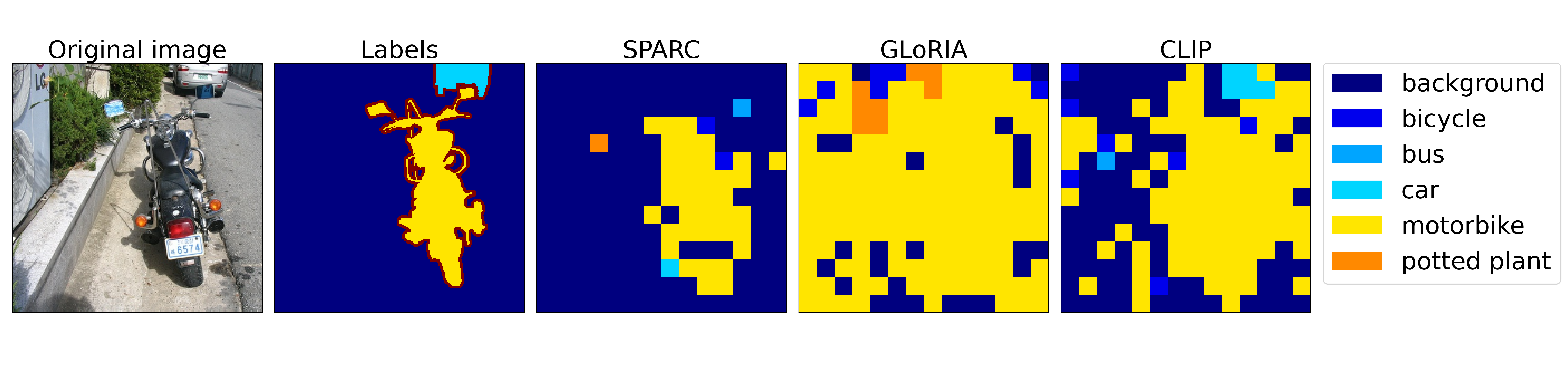} }}%
    	\vspace{-3mm}
    	\quad
    	\subfloat{{\includegraphics[height=2.75cm]{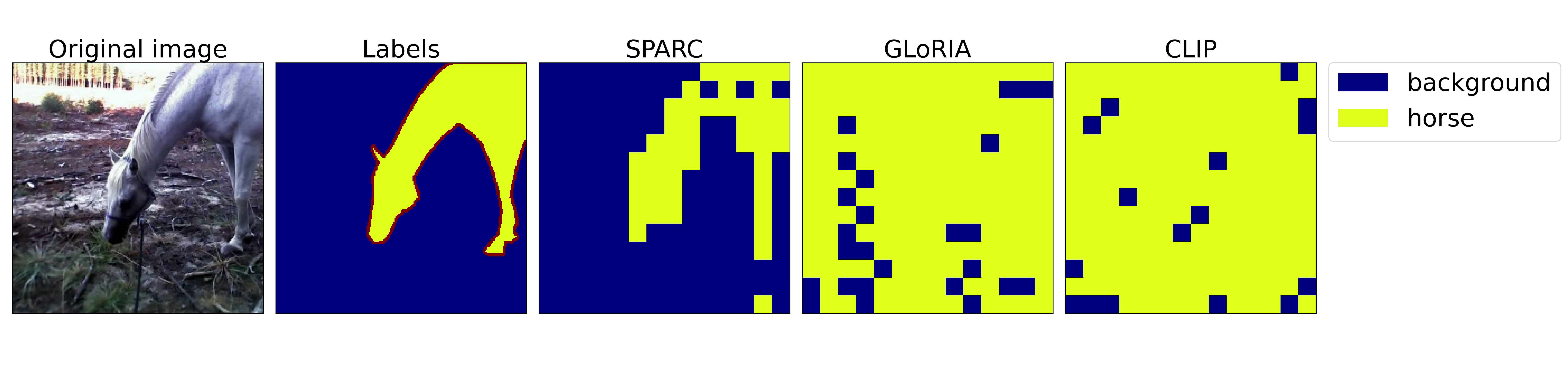} }}%
    	\caption{Qualitative results for zero-shot segmentation on Pascal VOC dataset. We illustrate the original image, pixel-level ground-truth labels and the the patch-level segmentation masks obtained from SPARC, GLoRIA and CLIP.}
    	\label{fig:semantic_segmentation}
    \end{figure}

\paragraph{Semantic Segmentation.}
Following related work \citep{pacl}, we also perform zero-shot segmentation \textit{given} a text label, i.e. we compute patch embeddings of a given image and calculate the cosine similarity of the patch embedding with the text embeddings of all the ground-truth classes \citep{pacl, ranasinghe2022perceptual}. We assign a matching class for each patch as the text that corresponds to the maximum cosine similarity of that patch. We then upsample the patches to match the resolution of the ground-truth segmentation and calculate for each class the Intersection over Union (IoU) between the predicted and ground-truth segmentations; we report the mean of the IoU scores over the classes present in the ground-truth image. More details about this evaluation can found in Appendix \ref{app.experiments}.
From Table~\ref{tab:segmentation} we see that SPARC strongly improves over other baselines, significantly surpassing the next best model by $+4.34$ mIoU on the PASCAL VOC~\citep{Everingham15} dataset and by $+1.2$ mIoU  on the PASCAL Context \citep{mottaghi_cvpr14} dataset.
We visualize the predicted segmentation masks on the PASCAL VOC dataset in Figure~\ref{fig:semantic_segmentation}. Whereas CLIP predicts the object to be present in many different parts of the image, SPARC achieves better object localization and predicts their shapes more accurately.

\subsection{SPARC backbones in vision language models}

\begin{wraptable}{r}{7.5cm}
\begin{center}
 \begin{adjustbox}{max width=\columnwidth}
    \begin{tabular}{lCCC}
         \toprule
         Method &  \text{MSCOCO} & \text{Flickr30k} \\
         \midrule
         \midrule
         CLIP & 24.3 & 12.9 \\
         SPARC (ours) & \textbf{25.3}  &  \textbf{13.6} \\
         \bottomrule
    \end{tabular}
    \end{adjustbox}
    \end{center}
    \caption{CIDEr score evaluating captioning performance of different vision backbones in a Flamingo-style \citep{flamingo} model.}
    \label{tab:flamingo}
\end{wraptable}

Vision backbones trained contrastively from image-text paired data are often frozen and used in foundational vision-language models (VLMs) such as Flamingo \citep{flamingo}. To understand whether the fine-grained performance improvements obtained from SPARC translate to better captioning performance in VLMs, we perform experiments where we compare using a CLIP backbone vs. a SPARC backbone in a Flamingo-style architecture~\citep{flamingo}. For this, we freeze the ViT-B/16 vision models trained with CLIP and SPARC and pair them with a frozen 400M parameter (pre-trained) language model. On top of the frozen vision and language backbones, we train Perceiver Resampler cross-attention layers \citep{flamingo} to produce free-form text as output. More details about the training set-up can be found in Appendix \ref{app.experiments}. We evaluate the models on captioning tasks on MSCOCO and Flickr30k datasets and we report results in Table \ref{tab:flamingo}.

\subsection{Ablations}

To assess the benefits of the different components in SPARC on performance, we perform the following two ablations: removing the sparsity on the similarity matrix and using softmax instead to compute the alignment weights for grouping the patch embeddings. From the results in Table \ref{tab:ablations} on both fine-grained (MSCOCO retrieval) and coarse-grained (ImageNet zero-shot classification) tasks we notice that both components play a significant role in the model's performance. In particular, using softmax results in the highest decrease in performance. See Appendix \ref{app.softmax} for a detailed discussion of the problems with using softmax to compute the alignment weights.

\begin{table}[h]
\begin{center}
 \begin{adjustbox}{max width=\columnwidth}
\begin{tabular}{lCCCCCCCCCCCC}
\toprule
  &  \multicolumn{2}{c}{MSCOCO (i2t)} & \multicolumn{2}{c}{MSCOCO (t2i)} & \multicolumn{1}{c}{ImageNet}  \\
    & \text{R@1} & \text{R@5} & \text{R@1} & \text{R@5} & \text{Top-1 acc.} \\
\midrule
\midrule
 SPARC & {\bf 57.6} &	{\bf 81.2} & {\bf 43.0} & {\bf 68.6} & {\bf 72.6}  \\
  - no sparsity & 56.1	& 80.7 & 42.4 & 68.2 & 72.1 \\ 
  - softmax & 55.2	& 79.8 & 41.6 &	67.5 & 70.6 \\
\bottomrule
\end{tabular}

 \end{adjustbox}
\caption{Ablations for the ViT-B/16 SPARC model on the MSCOCO image-to-text (i2t) and text-to-image (t2i) retrieval and zero-shot classification on ImageNet. }
\label{tab:ablations}
\end{center}

\end{table}

\subsection{Memory consumption and FLOPS} \label{sec.flops}

\begin{figure*}[h]%
    \centering
    \subfloat[\centering ]{{\includegraphics[width=7.5cm]{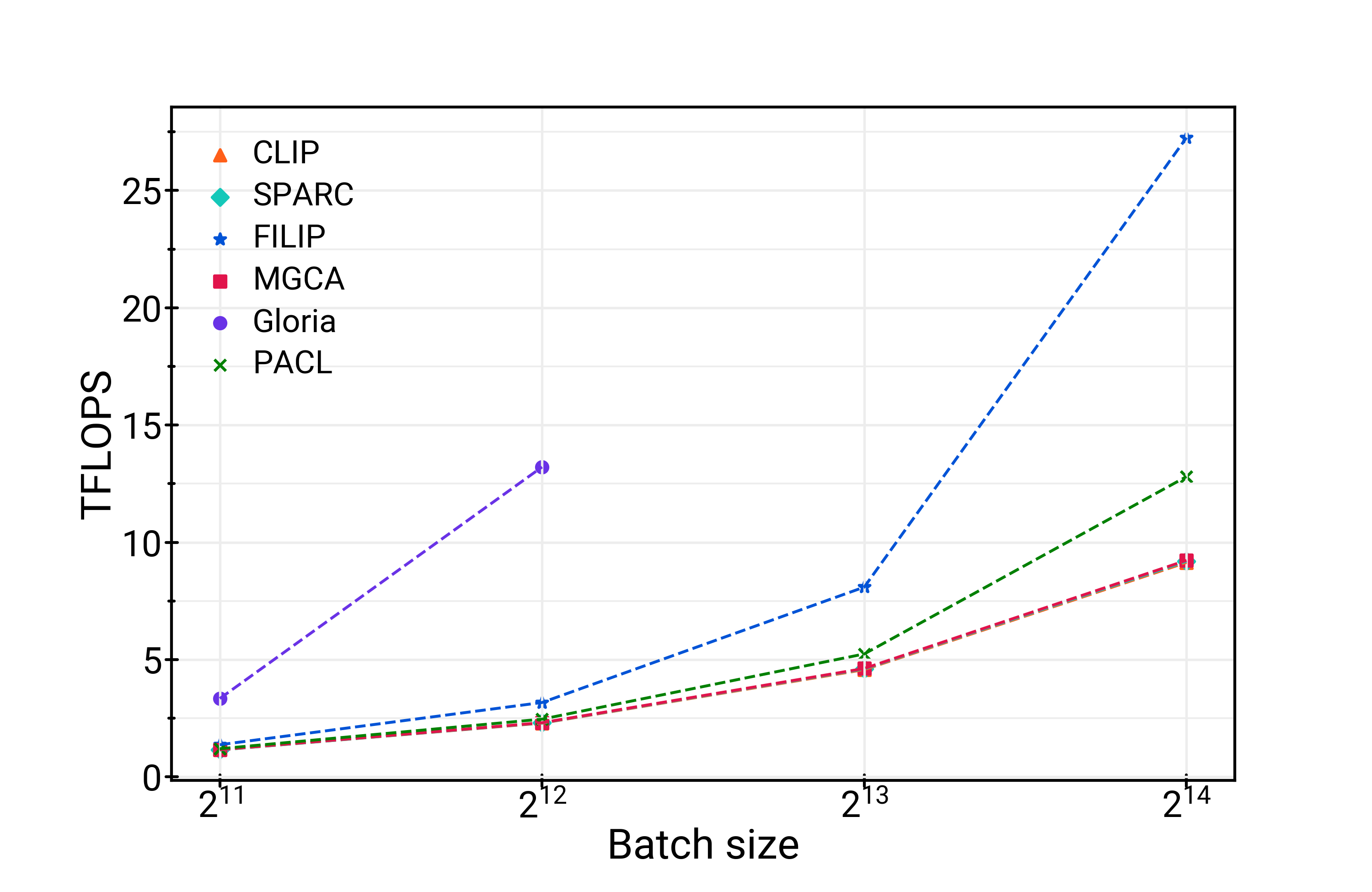} }}%
    \qquad
    \subfloat[\centering ]{{\includegraphics[width=7.5cm]{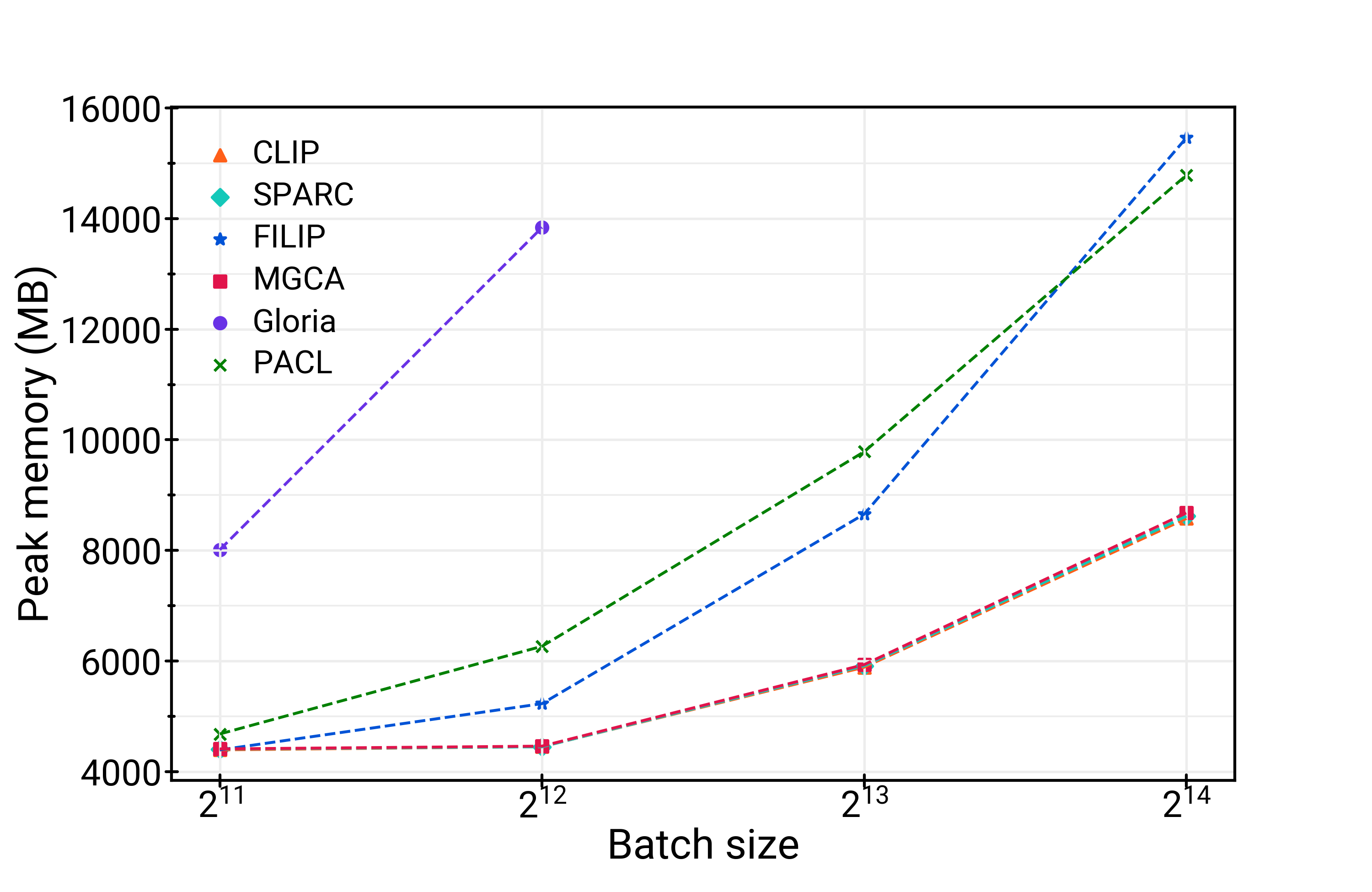} }}%
    \qquad
     \subfloat[\centering ]{{\includegraphics[width=7.5cm]{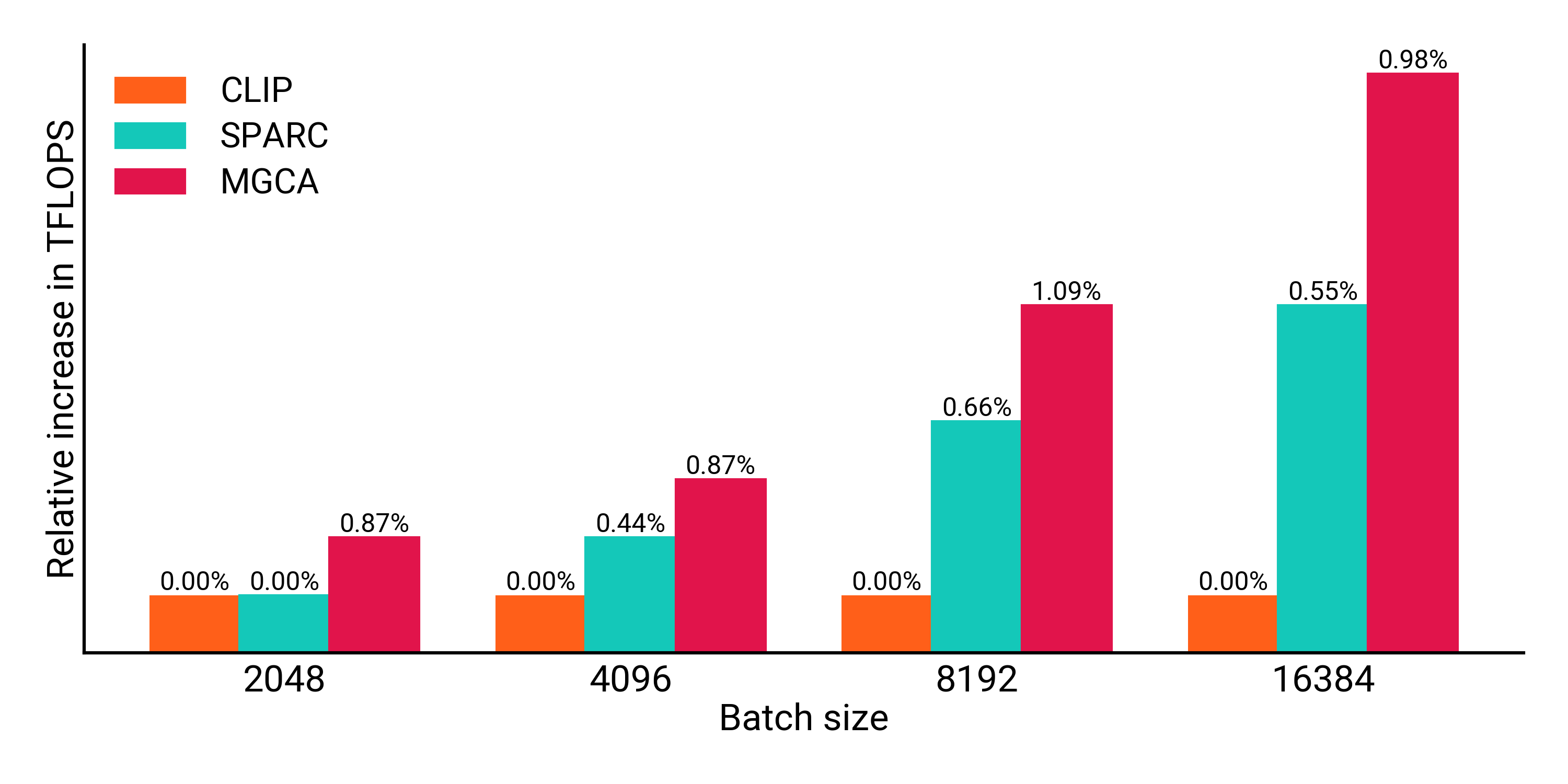} }}%
    \qquad
     \subfloat[\centering ]{{\includegraphics[width=7.5cm]{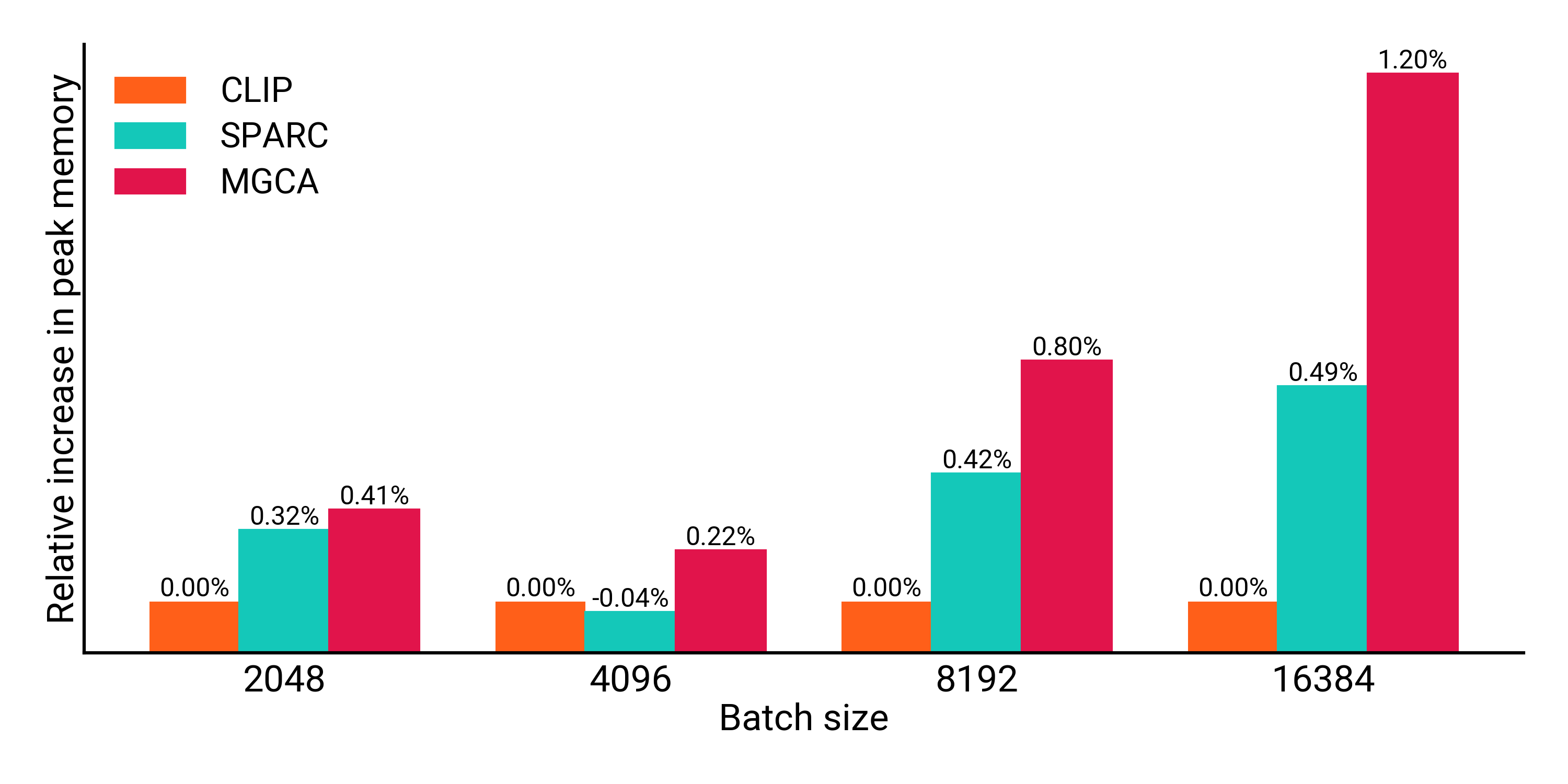} }}%
    \qquad
    \caption{TFLOPS (a) and Peak Memory (b) used by all methods. Relative increase in TFLOPS (c) and Peak memory (d) when comparing SPARC and MGCA to CLIP.}%
    \label{fig:flops_memory}%
\end{figure*}

To understand the computational and memory efficiency of the different methods, we also compute the FLOPS and peak memory usage for one update step for different batch size. Note that all methods are trained on 256 TPUs. In Figure \ref{fig:flops_memory}~(a) we show the teraFLOPS (TFLOPS) and in Figure \ref{fig:flops_memory}~(b) the peak memory usage (in MB) of the different methods for one update step when varying the batch size (B) from 2048 to 16384. Notice that GLoRIA \citep{gloria} is as memory intensive at batch size 4096 as the other methods (e.g. CLIP) at batch size 16384. Thus, due to device constraints, we were only able to train GLoRIA with batch size 4096. Moreover, notice that for FILIP the TFLOPS used for one update step increases by more than 200\% between B=8196 and B=16384, as opposed to the 100\% increase for CLIP, SPARC and MGCA. In addition, for B=16384, both FILIP and PACL have ~2x peak memory compared to CLIP, SPARC and MGCA. On the other hand, note that CLIP, SPARC and MGCA use the same order of magnitude of FLOPS and memory. To further highlight the differences between them, we plot the relative increase in TFLOPS in Figure \ref{fig:flops_memory}~(c) and the relative increase in peak memory in Figure \ref{fig:flops_memory}~(c) of SPARC and MGCA with respect to CLIP. Notice that for  B=16384, i.e. the batch size we use for our experiments, the relative increase in TFLOPS and peak memory for SPARC is almost half the one for MGCA. We provide detailed numbers for the FLOPS (in TFLOPS) and of the Peak Memory (in MB) in Appendix \ref{apx.flops}.

\newpage
\section{Discussion}

In this work we proposed a novel method Sparse Fine-grained Contrastive Alignment (SPARC) for fine-grained vision-language pretraining. 
SPARC simultaneously learns information at different levels of granularity by contrasting both image-level and caption-level embeddings and token and patch embeddings. 
SPARC learns to group patches based on similarity to tokens and contrast the resulting language-grounded patch embeddings with token embeddings.  
Unlike previous work this comparison is done within individual image-text pairs and does not require the computationally and memory expensive comparison of all patches and tokens within the full batch.
Through extensive experimental evaluation we show that SPARC improves performance both on image-level tasks like classification and retrieval and more fine-grained tasks like object detection and segmentation that require localization. Moreover, SPARC improves model faithfulness and 

While the simple sparsification of the similarity matrix in SPARC already improves performance, we believe that exploring different approaches to sparsification and learning patch groupings could lead to even more informative representations. 
Moreover, given that SPARC learns patch groupings based on the associated caption, exploring pretraining data with highly descriptive captions is another interesting line of future work.
Also, leveraging bounding boxes and segmentation masks (in addition to image-text pairs) would facilitate learning patch groupings and improve learning efficiency since the similarity matrix could be pre-sparsified according to these signals.
Another interesting avenue of future work is further exploring how SPARC encoders perform as part of multimodal foundational models like Flamingo~\citep{flamingo}, BLIP~\citep{blip} and PALI~\citep{pali}.

\bibliographystyle{abbrvnat}
\nobibliography*
\bibliography{main}

\begin{thebibliography}{59}
\providecommand{\natexlab}[1]{#1}
\providecommand{\url}[1]{\texttt{#1}}
\expandafter\ifx\csname urlstyle\endcsname\relax
  \providecommand{\doi}[1]{doi: #1}\else
  \providecommand{\doi}{doi: \begingroup \urlstyle{rm}\Url}\fi

\bibitem[Adlakha et~al.(2023)Adlakha, BehnamGhader, Lu, Meade, and
  Reddy]{adlakha2023evaluating}
V.~Adlakha, P.~BehnamGhader, X.~H. Lu, N.~Meade, and S.~Reddy.
\newblock Evaluating correctness and faithfulness of instruction-following
  models for question answering.
\newblock \emph{arXiv preprint arXiv:2307.16877}, 2023.

\bibitem[Alayrac et~al.(2022)Alayrac, Donahue, Luc, Miech, Barr, Hasson, Lenc,
  Mensch, Millican, Reynolds, et~al.]{flamingo}
J.-B. Alayrac, J.~Donahue, P.~Luc, A.~Miech, I.~Barr, Y.~Hasson, K.~Lenc,
  A.~Mensch, K.~Millican, M.~Reynolds, et~al.
\newblock Flamingo: a visual language model for few-shot learning.
\newblock \emph{Advances in Neural Information Processing Systems},
  35:\penalty0 23716--23736, 2022.

\bibitem[Alsentzer et~al.(2019)Alsentzer, Murphy, Boag, Weng, Jin, Naumann, and
  McDermott]{alsentzer2019publicly}
E.~Alsentzer, J.~R. Murphy, W.~Boag, W.-H. Weng, D.~Jin, T.~Naumann, and
  M.~McDermott.
\newblock Publicly available clinical bert embeddings.
\newblock \emph{arXiv preprint arXiv:1904.03323}, 2019.

\bibitem[Chen et~al.(2022)Chen, Wang, Changpinyo, Piergiovanni, Padlewski,
  Salz, Goodman, Grycner, Mustafa, Beyer, et~al.]{pali}
X.~Chen, X.~Wang, S.~Changpinyo, A.~Piergiovanni, P.~Padlewski, D.~Salz,
  S.~Goodman, A.~Grycner, B.~Mustafa, L.~Beyer, et~al.
\newblock Pali: A jointly-scaled multilingual language-image model.
\newblock \emph{arXiv preprint arXiv:2209.06794}, 2022.

\bibitem[Dawidowicz et~al.(2023)Dawidowicz, Hirsch, and
  Tal]{Dawidowicz2023LIMITRLL}
G.~Dawidowicz, E.~Hirsch, and A.~Tal.
\newblock Limitr: Leveraging local information for medical image-text
  representation.
\newblock \emph{ArXiv}, abs/2303.11755, 2023.
\newblock URL \url{https://api.semanticscholar.org/CorpusID:257636659}.

\bibitem[Dosovitskiy et~al.(2020)Dosovitskiy, Beyer, Kolesnikov, Weissenborn,
  Zhai, Unterthiner, Dehghani, Minderer, Heigold, Gelly, et~al.]{vit}
A.~Dosovitskiy, L.~Beyer, A.~Kolesnikov, D.~Weissenborn, X.~Zhai,
  T.~Unterthiner, M.~Dehghani, M.~Minderer, G.~Heigold, S.~Gelly, et~al.
\newblock An image is worth 16x16 words: Transformers for image recognition at
  scale.
\newblock \emph{arXiv preprint arXiv:2010.11929}, 2020.

\bibitem[Elfadel and Wyatt~Jr(1993)]{elfadel1993softmax}
I.~M. Elfadel and J.~L. Wyatt~Jr.
\newblock The" softmax" nonlinearity: Derivation using statistical mechanics
  and useful properties as a multiterminal analog circuit element.
\newblock \emph{Advances in neural information processing systems}, 6, 1993.

\bibitem[Everingham et~al.(2015)Everingham, Eslami, Van~Gool, Williams, Winn,
  and Zisserman]{Everingham15}
M.~Everingham, S.~M.~A. Eslami, L.~Van~Gool, C.~K.~I. Williams, J.~Winn, and
  A.~Zisserman.
\newblock The pascal visual object classes challenge: A retrospective.
\newblock \emph{International Journal of Computer Vision}, 111\penalty0
  (1):\penalty0 98--136, Jan. 2015.

\bibitem[Geng et~al.(2023)Geng, Yuan, Tian, Chen, and Zhang]{geng2023hiclip}
S.~Geng, J.~Yuan, Y.~Tian, Y.~Chen, and Y.~Zhang.
\newblock Hiclip: Contrastive language-image pretraining with hierarchy-aware
  attention.
\newblock \emph{arXiv preprint arXiv:2303.02995}, 2023.

\bibitem[Gupta et~al.(2019)Gupta, Dollar, and Girshick]{gupta2019lvis}
A.~Gupta, P.~Dollar, and R.~Girshick.
\newblock Lvis: A dataset for large vocabulary instance segmentation.
\newblock In \emph{Proceedings of the IEEE/CVF conference on computer vision
  and pattern recognition}, pages 5356--5364, 2019.

\bibitem[Hendrycks and Dietterich(2019)]{hendrycks2019benchmarking}
D.~Hendrycks and T.~Dietterich.
\newblock Benchmarking neural network robustness to common corruptions and
  perturbations.
\newblock \emph{arXiv preprint arXiv:1903.12261}, 2019.

\bibitem[Hendrycks et~al.(2019)Hendrycks, Zhao, Basart, Steinhardt, and
  Song]{hendrycks2019natural}
D.~Hendrycks, K.~Zhao, S.~Basart, J.~Steinhardt, and D.~Song.
\newblock Natural adversarial examples.(2019).
\newblock \emph{arXiv preprint cs.LG/1907.07174}, 5\penalty0 (6), 2019.

\bibitem[Hendrycks et~al.(2021)Hendrycks, Basart, Mu, Kadavath, Wang, Dorundo,
  Desai, Zhu, Parajuli, Guo, et~al.]{hendrycks2021many}
D.~Hendrycks, S.~Basart, N.~Mu, S.~Kadavath, F.~Wang, E.~Dorundo, R.~Desai,
  T.~Zhu, S.~Parajuli, M.~Guo, et~al.
\newblock The many faces of robustness: A critical analysis of
  out-of-distribution generalization.
\newblock In \emph{Proceedings of the IEEE/CVF International Conference on
  Computer Vision}, pages 8340--8349, 2021.

\bibitem[Hoffmann et~al.(2023)Hoffmann, Schrodi, Behrmann, Fischer, and
  Brox]{hoffmann2023softmax}
D.~T. Hoffmann, S.~Schrodi, N.~Behrmann, V.~Fischer, and T.~Brox.
\newblock Eureka-moments in transformers: Multi-step tasks reveal softmax
  induced optimization problems.
\newblock \emph{arXiv preprint arXiv:2310.12956}, 2023.

\bibitem[Huang et~al.(2021)Huang, Shen, Lungren, and Yeung]{gloria}
S.-C. Huang, L.~Shen, M.~P. Lungren, and S.~Yeung.
\newblock Gloria: A multimodal global-local representation learning framework
  for label-efficient medical image recognition.
\newblock In \emph{Proceedings of the IEEE/CVF International Conference on
  Computer Vision}, pages 3942--3951, 2021.

\bibitem[Ji et~al.(2023)Ji, Lee, Frieske, Yu, Su, Xu, Ishii, Bang, Madotto, and
  Fung]{ji2023survey}
Z.~Ji, N.~Lee, R.~Frieske, T.~Yu, D.~Su, Y.~Xu, E.~Ishii, Y.~J. Bang,
  A.~Madotto, and P.~Fung.
\newblock Survey of hallucination in natural language generation.
\newblock \emph{ACM Computing Surveys}, 55\penalty0 (12):\penalty0 1--38, 2023.

\bibitem[Jia et~al.(2021)Jia, Yang, Xia, Chen, Parekh, Pham, Le, Sung, Li, and
  Duerig]{align}
C.~Jia, Y.~Yang, Y.~Xia, Y.-T. Chen, Z.~Parekh, H.~Pham, Q.~Le, Y.-H. Sung,
  Z.~Li, and T.~Duerig.
\newblock Scaling up visual and vision-language representation learning with
  noisy text supervision.
\newblock In \emph{International conference on machine learning}, pages
  4904--4916. PMLR, 2021.

\bibitem[Krishna et~al.(2017)Krishna, Zhu, Groth, Johnson, Hata, Kravitz, Chen,
  Kalantidis, Li, Shamma, et~al.]{krishna2017visual}
R.~Krishna, Y.~Zhu, O.~Groth, J.~Johnson, K.~Hata, J.~Kravitz, S.~Chen,
  Y.~Kalantidis, L.-J. Li, D.~A. Shamma, et~al.
\newblock Visual genome: Connecting language and vision using crowdsourced
  dense image annotations.
\newblock \emph{International journal of computer vision}, 123:\penalty0
  32--73, 2017.

\bibitem[Krojer et~al.(2022)Krojer, Adlakha, Vineet, Goyal, Ponti, and
  Reddy]{krojer2022image}
B.~Krojer, V.~Adlakha, V.~Vineet, Y.~Goyal, E.~Ponti, and S.~Reddy.
\newblock Image retrieval from contextual descriptions.
\newblock \emph{arXiv preprint arXiv:2203.15867}, 2022.

\bibitem[Kudo and Richardson(2018)]{kudo2018sentencepiece}
T.~Kudo and J.~Richardson.
\newblock Sentencepiece: A simple and language independent subword tokenizer
  and detokenizer for neural text processing.
\newblock \emph{arXiv preprint arXiv:1808.06226}, 2018.

\bibitem[Kuznetsova et~al.(2020)Kuznetsova, Rom, Alldrin, Uijlings, Krasin,
  Pont-Tuset, Kamali, Popov, Malloci, Kolesnikov, et~al.]{kuznetsova2020open}
A.~Kuznetsova, H.~Rom, N.~Alldrin, J.~Uijlings, I.~Krasin, J.~Pont-Tuset,
  S.~Kamali, S.~Popov, M.~Malloci, A.~Kolesnikov, et~al.
\newblock The open images dataset v4: Unified image classification, object
  detection, and visual relationship detection at scale.
\newblock \emph{International Journal of Computer Vision}, 128\penalty0
  (7):\penalty0 1956--1981, 2020.

\bibitem[Li et~al.(2021)Li, Selvaraju, Gotmare, Joty, Xiong, and Hoi]{albef}
J.~Li, R.~Selvaraju, A.~Gotmare, S.~Joty, C.~Xiong, and S.~C.~H. Hoi.
\newblock Align before fuse: Vision and language representation learning with
  momentum distillation.
\newblock \emph{Advances in neural information processing systems},
  34:\penalty0 9694--9705, 2021.

\bibitem[Li et~al.(2022{\natexlab{a}})Li, Li, Xiong, and Hoi]{blip}
J.~Li, D.~Li, C.~Xiong, and S.~Hoi.
\newblock Blip: Bootstrapping language-image pre-training for unified
  vision-language understanding and generation.
\newblock In \emph{International Conference on Machine Learning}, pages
  12888--12900. PMLR, 2022{\natexlab{a}}.

\bibitem[Li et~al.(2022{\natexlab{b}})Li, Zhang, Zhang, Yang, Li, Zhong, Wang,
  Yuan, Zhang, Hwang, et~al.]{li2022grounded}
L.~H. Li, P.~Zhang, H.~Zhang, J.~Yang, C.~Li, Y.~Zhong, L.~Wang, L.~Yuan,
  L.~Zhang, J.-N. Hwang, et~al.
\newblock Grounded language-image pre-training.
\newblock In \emph{Proceedings of the IEEE/CVF Conference on Computer Vision
  and Pattern Recognition}, pages 10965--10975, 2022{\natexlab{b}}.

\bibitem[Lin et~al.(2014)Lin, Maire, Belongie, Hays, Perona, Ramanan,
  Doll{\'a}r, and Zitnick]{lin2014microsoft}
T.-Y. Lin, M.~Maire, S.~Belongie, J.~Hays, P.~Perona, D.~Ramanan,
  P.~Doll{\'a}r, and C.~L. Zitnick.
\newblock Microsoft coco: Common objects in context.
\newblock In \emph{Computer Vision--ECCV 2014: 13th European Conference,
  Zurich, Switzerland, September 6-12, 2014, Proceedings, Part V 13}, pages
  740--755. Springer, 2014.

\bibitem[Loshchilov and Hutter(2017)]{loshchilov2017decoupled}
I.~Loshchilov and F.~Hutter.
\newblock Decoupled weight decay regularization.
\newblock \emph{arXiv preprint arXiv:1711.05101}, 2017.

\bibitem[Minderer et~al.(2022)Minderer, Gritsenko, Stone, Neumann, Weissenborn,
  Dosovitskiy, Mahendran, Arnab, Dehghani, Shen, et~al.]{minderer2022simple}
M.~Minderer, A.~Gritsenko, A.~Stone, M.~Neumann, D.~Weissenborn,
  A.~Dosovitskiy, A.~Mahendran, A.~Arnab, M.~Dehghani, Z.~Shen, et~al.
\newblock Simple open-vocabulary object detection.
\newblock In \emph{European Conference on Computer Vision}, pages 728--755.
  Springer, 2022.

\bibitem[Mottaghi et~al.(2014)Mottaghi, Chen, Liu, Cho, Lee, Fidler, Urtasun,
  and Yuille]{mottaghi_cvpr14}
R.~Mottaghi, X.~Chen, X.~Liu, N.-G. Cho, S.-W. Lee, S.~Fidler, R.~Urtasun, and
  A.~Yuille.
\newblock The role of context for object detection and semantic segmentation in
  the wild.
\newblock In \emph{IEEE Conference on Computer Vision and Pattern Recognition
  (CVPR)}, 2014.

\bibitem[Mukhoti et~al.(2023)Mukhoti, Lin, Poursaeed, Wang, Shah, Torr, and
  Lim]{pacl}
J.~Mukhoti, T.-Y. Lin, O.~Poursaeed, R.~Wang, A.~Shah, P.~H. Torr, and S.-N.
  Lim.
\newblock Open vocabulary semantic segmentation with patch aligned contrastive
  learning.
\newblock In \emph{Proceedings of the IEEE/CVF Conference on Computer Vision
  and Pattern Recognition}, pages 19413--19423, 2023.

\bibitem[Paiss et~al.(2023)Paiss, Ephrat, Tov, Zada, Mosseri, Irani, and
  Dekel]{paiss2023teaching}
R.~Paiss, A.~Ephrat, O.~Tov, S.~Zada, I.~Mosseri, M.~Irani, and T.~Dekel.
\newblock Teaching clip to count to ten.
\newblock \emph{arXiv preprint arXiv:2302.12066}, 2023.

\bibitem[Parcalabescu et~al.(2021)Parcalabescu, Cafagna, Muradjan, Frank,
  Calixto, and Gatt]{parcalabescu2021valse}
L.~Parcalabescu, M.~Cafagna, L.~Muradjan, A.~Frank, I.~Calixto, and A.~Gatt.
\newblock Valse: A task-independent benchmark for vision and language models
  centered on linguistic phenomena.
\newblock \emph{arXiv preprint arXiv:2112.07566}, 2021.

\bibitem[Peterson and S\"{o}derberg(1989)]{Peterson89}
C.~Peterson and B.~S\"{o}derberg.
\newblock A new method for mapping optimization problems onto neural networks.
\newblock \emph{International Journal of Neural Systems}, 01\penalty0
  (01):\penalty0 3--22, 1989.

\bibitem[Plummer et~al.(2015)Plummer, Wang, Cervantes, Caicedo, Hockenmaier,
  and Lazebnik]{plummer2015flickr30k}
B.~A. Plummer, L.~Wang, C.~M. Cervantes, J.~C. Caicedo, J.~Hockenmaier, and
  S.~Lazebnik.
\newblock Flickr30k entities: Collecting region-to-phrase correspondences for
  richer image-to-sentence models.
\newblock In \emph{Proceedings of the IEEE international conference on computer
  vision}, pages 2641--2649, 2015.

\bibitem[Radford et~al.(2021)Radford, Kim, Hallacy, Ramesh, Goh, Agarwal,
  Sastry, Askell, Mishkin, Clark, et~al.]{clip}
A.~Radford, J.~W. Kim, C.~Hallacy, A.~Ramesh, G.~Goh, S.~Agarwal, G.~Sastry,
  A.~Askell, P.~Mishkin, J.~Clark, et~al.
\newblock Learning transferable visual models from natural language
  supervision.
\newblock In \emph{International conference on machine learning}, pages
  8748--8763. PMLR, 2021.

\bibitem[Ranasinghe et~al.(2022)Ranasinghe, McKinzie, Ravi, Yang, Toshev, and
  Shlens]{ranasinghe2022perceptual}
K.~Ranasinghe, B.~McKinzie, S.~Ravi, Y.~Yang, A.~Toshev, and J.~Shlens.
\newblock Perceptual grouping in vision-language models.
\newblock \emph{arXiv preprint arXiv:2210.09996}, 2022.

\bibitem[Ranasinghe et~al.(2023)Ranasinghe, McKinzie, Ravi, Yang, Toshev, and
  Shlens]{Ranasinghe2022PerceptualGI}
K.~Ranasinghe, B.~McKinzie, S.~Ravi, Y.~Yang, A.~Toshev, and J.~Shlens.
\newblock Perceptual grouping in contrastive vision-language models.
\newblock In \emph{Proceedings of the IEEE/CVF International Conference on
  Computer Vision}, pages 5571--5584, 2023.

\bibitem[Razumovskaia et~al.(2023)Razumovskaia, Vulić, Marković, Cichy,
  Zheng, Wen, and Budzianowski]{razumovskaia2023textitdial}
E.~Razumovskaia, I.~Vulić, P.~Marković, T.~Cichy, Q.~Zheng, T.-H. Wen, and
  P.~Budzianowski.
\newblock $\textit{Dial BeInfo for Faithfulness}$: Improving factuality of
  information-seeking dialogue via behavioural fine-tuning, 2023.

\bibitem[Recht et~al.(2019)Recht, Roelofs, Schmidt, and
  Shankar]{recht2019imagenet}
B.~Recht, R.~Roelofs, L.~Schmidt, and V.~Shankar.
\newblock Do imagenet classifiers generalize to imagenet?
\newblock In \emph{International conference on machine learning}, pages
  5389--5400. PMLR, 2019.

\bibitem[Ren et~al.(2015)Ren, He, Girshick, and Sun]{ren2015faster}
S.~Ren, K.~He, R.~Girshick, and J.~Sun.
\newblock Faster r-cnn: Towards real-time object detection with region proposal
  networks.
\newblock \emph{Advances in neural information processing systems}, 28, 2015.

\bibitem[Russakovsky et~al.(2015)Russakovsky, Deng, Su, Krause, Satheesh, Ma,
  Huang, Karpathy, Khosla, Bernstein, et~al.]{imagenet}
O.~Russakovsky, J.~Deng, H.~Su, J.~Krause, S.~Satheesh, S.~Ma, Z.~Huang,
  A.~Karpathy, A.~Khosla, M.~Bernstein, et~al.
\newblock Imagenet large scale visual recognition challenge.
\newblock \emph{International journal of computer vision}, 115:\penalty0
  211--252, 2015.

\bibitem[Shao et~al.(2019)Shao, Li, Zhang, Peng, Yu, Zhang, Li, and
  Sun]{shao2019objects365}
S.~Shao, Z.~Li, T.~Zhang, C.~Peng, G.~Yu, X.~Zhang, J.~Li, and J.~Sun.
\newblock Objects365: A large-scale, high-quality dataset for object detection.
\newblock In \emph{Proceedings of the IEEE/CVF international conference on
  computer vision}, pages 8430--8439, 2019.

\bibitem[Shen et~al.(2023)Shen, Guo, Tan, Tang, Wang, and
  Bian]{shen2023softmax}
K.~Shen, J.~Guo, X.~Tan, S.~Tang, R.~Wang, and J.~Bian.
\newblock A study on relu and softmax in transformer.
\newblock \emph{arXiv preprint arXiv:2302.06461}, 2023.

\bibitem[Sun et~al.(2017)Sun, Shrivastava, Singh, and Gupta]{sun2017revisiting}
C.~Sun, A.~Shrivastava, S.~Singh, and A.~Gupta.
\newblock Revisiting unreasonable effectiveness of data in deep learning era.
\newblock In \emph{Proceedings of the IEEE international conference on computer
  vision}, pages 843--852, 2017.

\bibitem[Varma et~al.(2023)Varma, Delbrouck, Hooper, Chaudhari, and
  Langlotz]{varma2023villa}
M.~Varma, J.-B. Delbrouck, S.~Hooper, A.~Chaudhari, and C.~Langlotz.
\newblock Villa: Fine-grained vision-language representation learning from
  real-world data.
\newblock In \emph{Proceedings of the IEEE/CVF International Conference on
  Computer Vision}, pages 22225--22235, 2023.

\bibitem[Vaswani et~al.(2017)Vaswani, Shazeer, Parmar, Uszkoreit, Jones, Gomez,
  Kaiser, and Polosukhin]{transformer}
A.~Vaswani, N.~Shazeer, N.~Parmar, J.~Uszkoreit, L.~Jones, A.~N. Gomez,
  {\L}.~Kaiser, and I.~Polosukhin.
\newblock Attention is all you need.
\newblock \emph{Advances in neural information processing systems}, 30, 2017.

\bibitem[Wang et~al.(2022)Wang, Zhou, Wang, Vardhanabhuti, and Yu]{mgca}
F.~Wang, Y.~Zhou, S.~Wang, V.~Vardhanabhuti, and L.~Yu.
\newblock Multi-granularity cross-modal alignment for generalized medical
  visual representation learning.
\newblock \emph{Advances in Neural Information Processing Systems},
  35:\penalty0 33536--33549, 2022.

\bibitem[Wang et~al.(2019)Wang, Ge, Lipton, and Xing]{wang2019learning}
H.~Wang, S.~Ge, Z.~Lipton, and E.~P. Xing.
\newblock Learning robust global representations by penalizing local predictive
  power.
\newblock In \emph{Advances in Neural Information Processing Systems}, pages
  10506--10518, 2019.

\bibitem[Xu et~al.(2022)Xu, De~Mello, Liu, Byeon, Breuel, Kautz, and
  Wang]{xu2022groupvit}
J.~Xu, S.~De~Mello, S.~Liu, W.~Byeon, T.~Breuel, J.~Kautz, and X.~Wang.
\newblock Groupvit: Semantic segmentation emerges from text supervision.
\newblock In \emph{Proceedings of the IEEE/CVF Conference on Computer Vision
  and Pattern Recognition}, pages 18134--18144, 2022.

\bibitem[Xu et~al.(2023)Xu, Hou, Zhang, Feng, Wang, Qiao, and
  Xie]{xu2023learning}
J.~Xu, J.~Hou, Y.~Zhang, R.~Feng, Y.~Wang, Y.~Qiao, and W.~Xie.
\newblock Learning open-vocabulary semantic segmentation models from natural
  language supervision.
\newblock In \emph{Proceedings of the IEEE/CVF Conference on Computer Vision
  and Pattern Recognition}, pages 2935--2944, 2023.

\bibitem[Yang et~al.(2022)Yang, Duan, Tran, Xu, Chanda, Chen, Zeng, Chilimbi,
  and Huang]{yang2022vision}
J.~Yang, J.~Duan, S.~Tran, Y.~Xu, S.~Chanda, L.~Chen, B.~Zeng, T.~Chilimbi, and
  J.~Huang.
\newblock Vision-language pre-training with triple contrastive learning.
\newblock In \emph{Proceedings of the IEEE/CVF Conference on Computer Vision
  and Pattern Recognition}, pages 15671--15680, 2022.

\bibitem[Yao et~al.(2021)Yao, Huang, Hou, Lu, Niu, Xu, Liang, Li, Jiang, and
  Xu]{filip}
L.~Yao, R.~Huang, L.~Hou, G.~Lu, M.~Niu, H.~Xu, X.~Liang, Z.~Li, X.~Jiang, and
  C.~Xu.
\newblock Filip: Fine-grained interactive language-image pre-training.
\newblock \emph{arXiv preprint arXiv:2111.07783}, 2021.

\bibitem[Yu et~al.(2022)Yu, Wang, Vasudevan, Yeung, Seyedhosseini, and
  Wu]{coca}
J.~Yu, Z.~Wang, V.~Vasudevan, L.~Yeung, M.~Seyedhosseini, and Y.~Wu.
\newblock Coca: Contrastive captioners are image-text foundation models.
\newblock \emph{arXiv preprint arXiv:2205.01917}, 2022.

\bibitem[Yuksekgonul et~al.(2022)Yuksekgonul, Bianchi, Kalluri, Jurafsky, and
  Zou]{aro}
M.~Yuksekgonul, F.~Bianchi, P.~Kalluri, D.~Jurafsky, and J.~Zou.
\newblock When and why vision-language models behave like bag-of-words models,
  and what to do about it?
\newblock \emph{arXiv preprint arXiv:2210.01936}, 2022.

\bibitem[Zeng et~al.(2021)Zeng, Zhang, and Li]{xvlm}
Y.~Zeng, X.~Zhang, and H.~Li.
\newblock Multi-grained vision language pre-training: Aligning texts with
  visual concepts.
\newblock \emph{arXiv preprint arXiv:2111.08276}, 2021.

\bibitem[Zhai et~al.(2023{\natexlab{a}})Zhai, Likhomanenko, Littwin, Busbridge,
  Ramapuram, Zhang, Gu, and Susskind]{zhai2023softmax}
S.~Zhai, T.~Likhomanenko, E.~Littwin, D.~Busbridge, J.~Ramapuram, Y.~Zhang,
  J.~Gu, and J.~Susskind.
\newblock Stabilizing transformer training by preventing attention entropy
  collapse.
\newblock \emph{ICML}, 2023{\natexlab{a}}.

\bibitem[Zhai et~al.(2022)Zhai, Kolesnikov, Houlsby, and
  Beyer]{zhai2022scaling}
X.~Zhai, A.~Kolesnikov, N.~Houlsby, and L.~Beyer.
\newblock Scaling vision transformers.
\newblock In \emph{Proceedings of the IEEE/CVF Conference on Computer Vision
  and Pattern Recognition}, pages 12104--12113, 2022.

\bibitem[Zhai et~al.(2023{\natexlab{b}})Zhai, Mustafa, Kolesnikov, and
  Beyer]{sigclip}
X.~Zhai, B.~Mustafa, A.~Kolesnikov, and L.~Beyer.
\newblock Sigmoid loss for language image pre-training.
\newblock \emph{International Conference on Computer Vision},
  2023{\natexlab{b}}.

\bibitem[Zhong et~al.(2022)Zhong, Yang, Zhang, Li, Codella, Li, Zhou, Dai,
  Yuan, Li, et~al.]{zhong2022regionclip}
Y.~Zhong, J.~Yang, P.~Zhang, C.~Li, N.~Codella, L.~H. Li, L.~Zhou, X.~Dai,
  L.~Yuan, Y.~Li, et~al.
\newblock Regionclip: Region-based language-image pretraining.
\newblock In \emph{Proceedings of the IEEE/CVF Conference on Computer Vision
  and Pattern Recognition}, pages 16793--16803, 2022.

\bibitem[Zhou et~al.(2022)Zhou, Loy, and Dai]{zhou2022extract}
C.~Zhou, C.~C. Loy, and B.~Dai.
\newblock Extract free dense labels from clip.
\newblock In \emph{European Conference on Computer Vision}, pages 696--712.
  Springer, 2022.

\end{thebibliography}

\clearpage
\appendix

\section{Problems with using softmax for obtaining alignment weights} \label{app.softmax}

$Softmax$ is ubiquitously used to normalise activations that should or could be interpreted as probabilities, as it is for example the case of attention/alignmnet weights. One potential reason behind this choice is the dominating practice of using $softmax$ as the output activation function for classification tasks, being the canonical link function for multinomial outputs. Another appealing property is that it acts as a differentiable $max$-operator, allowing for a natural interpretation of \emph{selecting} one class out of multiple.

However, $softmax$ can be problematic from a gradient flow perspective~\citep{shen2023softmax,zhai2023softmax,hoffmann2023softmax}, and in this section we will expand this observation and the implications it might have on our specific use case. Also, intuitively from its role as a soften $max$ operator, softmax prefers to converge to peaky uni-modal distribution, selecting one out of $k$, and is less likely to represent multi-modal distributions. This is due to how gradients flow through the activation, leading to winner-takes-all dynamics~\citep{elfadel1993softmax,Peterson89} that ensures the peakyness and unimodality of the distribution represented. 

 If we assume $a(\hbf) = softmax(\hbf)$\footnote{By abuse of notation, we will use $\abf \in \mathcal{R}^k$, where $\abf = a(\hbf)$ and use $\abf_i$ for the $i$-th dimension of vector $\abf$}, for some $\hbf \in \mathcal{R}^k$, then we can write the derivative as 

\begin{equation}
\label{eq:diffsm}
    \frac{\partial \abf_i}{\partial \hbf_j} = 
            \left\{ \begin{array}{cc} \abf_i - \abf_i^2 & \text{iff } i = j \\ -\abf_i \abf_j & \text{otherwise} \end{array}\right.
\end{equation}

Assume we have some loss $L$ which is a function of $\sum_i \abf_i \Vbf_i$, i.e. some values $\Vbf_i \in \mathcal{R}^n$ that have been summarised using attention weights $\abf_i$. 

\paragraph{$\mathbf{Softmax}$ gradients vanish at initialisation.} Assume we have a large number of patches or tokens we want to attend over. In our notation, $k \gg 0$. At initialisation, all preactivation entries $\hbf_i$ will be small numbers of similar magnitude. The attention weights will be uniformally distributed over the $k$ patches, leading to $\abf_i \approx \frac{1}{k} \ll 1, \forall i$. 
Due to the weights being almost uniformally distributed, different observation will lead to randomly selecting a different patch. Therefore in expectation the gradient through the softmax on a particular token $i$ will be scaled by $\frac{1}{k^2}$ which will vanish very fast to 0 as $k$ grows. Note that in the rare scenario that the system picks the $i$-th element, the gradient becomes $\frac{1}{k}$ which also vanishes to 0 as $k$ grows. 
If we consider a very large $k$, this ensures that we have a plateau at initialization that might be hard to escape (or might take many updates to do so). 
See also \citep{hoffmann2023softmax} for a similar observation.

\paragraph{$\mathbf{Softmax}$ exhibits \emph{winner-takes-all} dynamics.} 
This has been understood and seen as a desirable property early on, see for example  \citep{Peterson89} and \citep{elfadel1993softmax}. One way to intuitively justify this behaviour is to think of the effect of applying the softmax operation multiple time (i.e. study the dynamics of a system whose transition function is just softmax). As shown in \citep{Peterson89} Fig. 5, the corners of the simplex act as attractors of this dynamical system, where from any initial condition, the system very quickly converges to one of the corners. 
This is caused by the dynamics of the gradients. When a particular weight is pushed up, all other weights are pushed down due to the normalisation. The amount by which the weight is pushed depends on its magnitude. So if a particular weight is larger and correlates positively with the desired behaviour, it will be pushed up proportionally more than other weights that correlate positively. Note that the particular form of the function (including the exponentiation) play a role in the form the gradients take, and removing the exponentiation will change the behaviour.  These types of dynamics,  have the downside of leading the distribution induced by the softmax to be unimodal. That is, softmax will act, as the name of the activation indicates, as a \emph{max} operator, preferring to learn a behaviour where it picks one out of $k$, rather than multiple equally relevant candidates.

\paragraph{$\mathbf{Softmax}$ saturates proportional to its certainty}
Assume $\exists i$ such that $\forall j, j\neq i$ we have $ a_i \gg a_j$. This implies that $1-a_i \to 0$ and $a_j < 1-a_i$. The gradient for the $i$-th position, according to equation~\ref{eq:diffsm}, will be $a_i (1-a_i)$ and will go to zero as linearly as $a_i$ approaches 1. The gradient for any other position $j$, will go to 0 at the same rate, as it will be roughly $a_j$ which is bounded from above from $1-a_i$. Note that a step of size $\Delta$ on $h$, due to the exponentiation and normalization of softmax, will make $a_i \to 1$ exponentially fast for constant change in $h$. 
\section{Additional related works} \label{app.related_works}

We further expand here the discussion on achieving  fine-grained understanding in vision-language models (VLMs) through additional losses and modules. 

In addition to the approaches described in Section \ref{sec:related_work}, another line of work involves proposes modifying the underlying vision transformer architecture to build modules that lead to a hierarchical grouping of image regions: e.g. GroupViT \citep{xu2022groupvit}, OVSegmentor \citep{xu2023learning}, HiCLIP \citep{geng2023hiclip}. While these methods propose architectural changes, the objective used for training still involves having a global contrastive loss. Conversely, in our work, we use the standard vision transformer architecture and propose instead changes to the training objective to achieve finegrained understanding. 

Moreover, note that several of these approaches \citep{xu2023learning} and the other methods who add a cross-modal encoder on top of the dual image-text encoder \citep{albef, yang2022vision} with captioning/masked language modelling losses start training from pre-trained text encoders and/or vision encoder. 

Similarly, \citep{Ranasinghe2022PerceptualGI} improve the semantic and spatial information in dual encoders trained contrastively by changing the patch embeddings aggregation methods from average pooling to max pooling and by starting training with both pre-trained vision and language encoders. In our work, we focus specifically on the set-up of training the dual encoders from scratch.

\clearpage
\newpage
\section{SPARC pseudo-code}
\label{app.pseudocode}

Listing~\ref{sup:pseudocode} provides JaX-alike pseudo-code for the SPARC objective detailing the construction of both the global and the local losses.

\noindent\begin{minipage}[t]{1.0\textwidth}
\begin{lstlisting}[language=Python, label={sup:pseudocode}, caption=Pseudo-code for SPARC.]
# Models:
#   vision_encoder
#   language_encoder
# Inputs:
#   image - (*@{\color{color_comments} [}@*)B, H, W, C(*@{\color{color_comments} ]}@*)
#   text - (*@{\color{color_comments}[@*)B, N(*@{\color{color_comments}]@*)
# Hyperparameters:
#   similarity_threshold
#   global_loss_weight
#   local_loss_weight
#   inverse_temperature

def pairwise_contrastive_loss((*@{\color{color_function_args}a@*), (*@{\color{color_function_args}b@*), (*@{\color{color_function_args}labels@*)):
    labels = eye(a.shape[0])
    logits_ab = dot(a * b.T) * inverse_temperature
    return softmax_cross_entropy(logits=logits_ab, labels=labels, reduction='mean')

def masked_pairwise_contrastive_loss((*@{\color{color_function_args}a@*), (*@{\color{color_function_args}b@*), (*@{\color{color_function_args}mask@*)):
    batch_size, seq_len, _ = a.shape[0]
    mask_logits = einshape('bnm->(*@{\color{color_strings}(@*)bn(*@{\color{color_strings})@*)m', 1.0 - mask, n=seq_len)
    labels = einshape('ns->(*@{\color{color_strings}(@*)bn(*@{\color{color_strings})@*)s', eye(a.shape[1]), b=batch_size)
    logits = einsum('bmd,bnd->bmn', a, b) * inverse_temperature
    logits = einshape('bnm->(*@{\color{color_strings}(@*)bn(*@{\color{color_strings})@*)m', logits)
    loss = softmax_cross_entropy(logits=logits - mask_logits * INF, labels=labels)
    loss = sum(loss * mask) / sum(mask)
    return loss

# ---------- GLOBAL LOSS ----------

# encoders include adapters
v_patch_embed = vision_encoder(image)
l_token_embed, language_mask = language_encoder(text)

v_embed = l2_normalize(mean(v_patch_embed, axis=1), axis=-1)
l_embed = l2_normalize(mean(l_token_embed, axis=1), axis=-1)

loss_vl = pairwise_contrastive_loss(v_embed, l_embed)
loss_lv = pairwise_contrastive_loss(l_embed, v_embed)

global_loss = 0.5 * (loss_vl + loss_lv)                                                             # (*@{\color{color_comments} (}@*)eq (*@{\color{color_comments} 1)}@*)

# ---------- LOCAL LOSS ----------

# similarity calculation
similarity = einsum('btd,bpd->btp', l_token_embed, v_patch_embed)

# min-max normalisation
similarity = (similarity - min(similarity, axis=-1)) /
             (max(similarity, axis=-1) - min(similarity, axis=-1))                                  # (*@{\color{color_comments} (}@*)eq 2(*@{\color{color_comments} )}@*)

# thresholding
similarity = where(similarity < similarity_threshold, 0.0, similarity)                              # (*@{\color{color_comments} (}@*)eq 3(*@{\color{color_comments} )}@*)

# alignment-weighting
v_align_weights = similarity / sum(similarity, axis=-1)                                             # (*@{\color{color_comments} (}@*)eq 4(*@{\color{color_comments} )}@*)
l_grouped_v_patch_embed = einsum('btp,bpd->btd', v_align_weights, v_patch_embed)                    # (*@{\color{color_comments} (}@*)eq 5(*@{\color{color_comments} )}@*)

l_grouped_v_patch_embed = l2_normalize(l_grouped_v_patch_embed, axis=-1)
l_token_embed = l2_normalize(l_token_embed, axis=-1)

loss_vl_local = masked_pairwise_contrastive_loss(l_grouped_v_patch_embed, l_token_embed, language_mask)
loss_lv_local = masked_pairwise_contrastive_loss(l_token_embed, l_grouped_v_patch_embed, language_mask)

local_loss = 0.5 * (loss_vl_local + loss_lv_local)                                                  # (*@{\color{color_comments} (}@*)eq 6(*@{\color{color_comments} )}@*)

# ---------- TOTAL (*@{\color{color_comments} (}@*)SPARC(*@{\color{color_comments} )}@*) LOSS ----------

loss = global_loss_weight * global_loss + local_loss_weight * local_loss                            # (*@{\color{color_comments} (}@*)eq 7(*@{\color{color_comments} )}@*)

\end{lstlisting}
\end{minipage}
\clearpage

\section{Experiments details} \label{app.experiments}

\subsection{Model architectures}

For the dual-encoder, we use the standard Vision Transformers (ViTs)~\citep{vit} as image encoders and Transformers~\citep{transformer} as text encoders. We perform experiments with ViT-B models with different patch sizes (ViT-B/32 and ViT-B/16) and a ViT-L model with patch size 14 (ViT-L/14). Thus, for the ViT-B image encoder, we use a model with 12 layers, 768 width and 12 attention heads, while for the ViT-L image encoder we use a model with 24 layers, 1024 width and 16 attention heads. For the language encoder, we use an architecture with 12 layers, 768 width and 12 attention heads. The linear adapters $g_v(\cdot)$ and $g_t(\cdot)$ project the vision and language embeddings respectively to a shared embedding space of dimensionality $512$.

\subsection{Datasets}

As described in Section \ref{sec:experiments}, we use the following datasets for pre-training: ALIGN~\citep{align}, JFT~\citep{sun2017revisiting, zhai2022scaling} and LTIP (Long Text \& Image Pairs)~\citep{flamingo}. Note that for JFT, where the images were semi-automatically annotated with a class-hierarchy of 30k labels, we flatten the hierarchical label structure and use all the assigned labels to describe the image. We use a multi-step training strategy where we alternate sampling batches from each of the 3 large datasets; the gradient updates are then performed by aggregating the gradients from computing the loss on one batch from each of the datasets.

\subsection{Baselines}

Our implementation of baselines follow the publicly available code (where available\footnote{GLoRIA: \url{https://github.com/marshuang80/gloria}, MGCA: \url{https://github.com/HKU-MedAI/MGCA}}) with a few minor differences we outline here. 

In the original MGCA implementation, token-wise cross-modal alignment (see Eqn. 5 in the original paper) uses the last-layer attention
weight from a visual token to the [CLS] token (averaged across multiple heads) to weight the loss terms for different visual tokens (and vice versa for language tokens). In our implementation, since we do not  use the [CLS] token but instead use average pooling to get the global language/vision embeddings, we omit this weighting operation. 

In the original GLoRIA implementation, language tokens are aggregated for each word to ensure that contrasted language embeddings refer to complete words (see Section 3.2.1 in the original paper); however, to ensure fair comparison, we do not have this additional aggregation operation, and instead use language tokens directly in local losses. Additionally, in our experiments we found that it is crucial to normalize the pairwise vision-language embedding similarities (see Eqn. 3 in the original paper) by $\sqrt{D}$ where $D$ is the embedding size. Without this normalization, we found training with  GLoRIA to be unstable. 
Moreover, recall that GLoRIA requires computing similarities between all token embeddings and all patch embeddings in the batch. This is memory expensive and it was not possible (due to device memory constraints) for batch sizes of 16348. Consequently, we used a batch size of 4096 for Gloria and trained the models for 800k steps (to match the number of examples seen by the other baseline). See discussion in Section \ref{apx.flops} for detailed computation of FLOPs and memory usage of GLoRIA.

For FILIP [50] we follow the original paper and implement token dropping for FILIP which the authors propose in order to reduce the large memory consumption of their method. In the original paper the authors comment on the training difficulty in the original paper (cf.  in the Appendix A.3.”...training is extremely unstable and the Nan loss easily happens.”). We observed similar training instability in our setup across a wide range of learning rates and weight decay parameters. This training instability leads to significant performance degradation compared to CLIP. We hypothesize that the non-standard additional tricks that FILIP uses such as image augmentations, backtranslation of captions and custom prompt ensembling could potentially improve training stability; note that we do not use these tricks in order to ensure a fair comparison across methods. Given FILIP's training instability, we conducted a number of additional experiments combining CLIP and FILIP in order to better understand the training instability. Below in Tables \ref{tab:app_classification} and \ref{tab:app_retrieval} we present these results -- as can be seen combining these two methods leads to some improvements on some benchmarks while some performance degradation on other benchmarks.

Finally, all methods in our paper use learned temperature parameters (instead of fixed temperatures as is done in the original MGCA and GLoRIA implementations) as our experiments showed that this significantly improved performance for all methods.

\begin{table*}[t!]
\begin{center}
\begin{tabular}{llCCCCCCCCCCCC}  %
\toprule
& Objective &  \text{IN} & \text{IN-V2 Th} & \text{IN-V2 MF} & \text{IN-V2 TI} & \text{IN-R} & \text{IN-C} & \text{IN-A}  & \text{IN-Sketch}  \\
\midrule
\midrule
\parbox[t]{2mm}{\multirow{4}{*}{\rotatebox[origin=c]{90}{ViT-B/32}}}  & CLIP & 66.7 &	66.2 &	58.9 &	71.5 &	63.2 & 42.6 &	15.1 &	51.7  \\   %
& FILIP  & 52.7 &	50.7	& 44.0	& 55.8 & 47.1 & 28.7 & 8.4 & 38.2  \\  %
& CLIP + FILIP & 66.5 &  65.8 & 58.2 & 71.1 & 63.0 & 42.3 & 15.1 & 51.3 	\\  %
& SPARC (ours) & {\bf 68.1} &	{\bf 67.0} & {\bf 59.7} &	{\bf 72.0} & {\bf 64.9} & {\bf 44.5} &	{\bf 16.7}	& {\bf 53.2}  \\   %
\midrule
\midrule
\parbox[t]{2mm}{\multirow{4}{*}{\rotatebox[origin=c]{90}{ViT-B/16}}}  
& CLIP & 71.6 &	70.9 &	63.7 &	74.8 & 71.1 &  {\bf 48.5}	& 32.2 &	56.8 \\  %
& FILIP  & 56.6 &	55.6	& 48.9	& 59.7 &  54.0  &   33.2   &   14.4   &  43.1  \\  %
& CLIP + FILIP & 71.8 & 70.5 & 63.4 & 74.4	& 70.6 & 47.8 & 32.0	& 56.2  \\ %
& SPARC (ours) &  {\bf 72.6}  &  {\bf 71.1}  &  {\bf 64.4}  &  {\bf 75.0}  &	{\bf 72.0} &	{\bf 48.5} & {\bf 33.8} &	{\bf 57.3} \\  %
\bottomrule
\end{tabular}
\end{center}
\caption{Top-1 accuracy (in \%) of zero-shot classification on ImageNet (IN) and its variants ImageNet-V2 Threshold (IN-V2 Th), ImageNet-V2 Matched Frequency (In-V2 MF), ImageNet-V2 Top Images (IN-V2 TI), ImageNet-R (IN-R), ImageNet-C (IN-C), ImageNet-Sketch (IN-Sketch). All methods have been trained on ALIGN, JFT, LTIP for the same number of training steps.}
\label{tab:app_classification}
\end{table*}

\begin{table*}[h]
\begin{center}
\begin{adjustbox}{max width=1.9\columnwidth}
\setlength\tabcolsep{3pt}
\begin{tabular}{llCCCCCCCCCCCCCCCC}  %
\toprule
 & &   \multicolumn{6}{c}{MSCOCO} & \multicolumn{6}{c}{Flickr30k} \\
 &  &  \multicolumn{3}{c}{image-to-text} & \multicolumn{3}{c}{text-to-image} & \multicolumn{3}{c} {image-to-text} & \multicolumn{3}{c} {text-to-image} \\
 &  Objective  & \text{R@1} & \text{R@5} & \text{R@10} & \text{R@1} & \text{R@5} & \text{R@10} & \text{R@1} & \text{R@5} & \text{R@10} & \text{R@1} & \text{R@5} & \text{R@10} \\
\midrule
\midrule
\parbox[t]{2mm}{\multirow{4}{*}{\rotatebox[origin=c]{90}{ViT-B/32}}}  & CLIP &  53.5 &	78.2 &	86.7 &	38.4 &	64.8 &	74.9 &	79.2 &	95.1 &	97.2 &	66.5 &	88.0 & {\bf 93.1}  \\
& FILIP  &  35.6 &	61.0 &	73.1	& 26.2 & 51.0 & 62.4 & 62.6 & 86.9 &	92.9 &	50.5 &	77.7 &	84.9	\\    %
& CLIP + FILIP & 52.0 &	77.0	& 85.6 &	37.8 &	64.4 &	74.5 &	81.2 &	95.4 &	97.1 &	66.8 &	87.7  &	92.3 \\  %
& SPARC (ours) & {\bf 55.0} &	{\bf 79.1} &	{\bf 87.3} &	{\bf 39.7} &	{\bf 65.9} &	{\bf 75.7} &	{\bf 82.5}	& {\bf 96.2} &	{\bf 97.6} &	{\bf 67.7} &	{\bf 88.2}  &	93.0 \\
\midrule
\midrule
\parbox[t]{2mm}{\multirow{4}{*}{\rotatebox[origin=c]{90}{ViT-B/16}}}  
& CLIP & 56.2 &	80.6 &	88.2 &	42.4 &	{\bf 68.6} &	78.3 &	84.0 &	96.1 &	98.2 &	71.6 &	90.3 & 94.1  \\
& FILIP  & 40.2 &	66.0 &	76.3 & 	29.5 & 	55.3	& 66.3	& 69.0	& 89.8	& 94.0	& 55.8	& 81.5 & 	87.9	 \\
& CLIP + FILIP &  54.9 & 79.0 &	87.4	& 41.3 & 67.7 &	77.5 &	82.7 & 97.0 &	98.4 &	71.1 & 90.5 &	94.7  \\ %
& SPARC (ours) & {\bf 57.6} &	{\bf 81.2} &	{\bf 88.5} &	{\bf 43.0} &	{\bf 68.6} &	{\bf 78.5} &	{\bf 84.4} &	{\bf 97.6} &	{\bf 98.7} &	{\bf 72.0} &	{\bf 91.2} &	{\bf 94.9} \\
\bottomrule
\end{tabular}
\end{adjustbox}
\end{center}
\caption{Results on zero-shot image-to-text and text-to-image retrieval on MSCOCO and Flickr30k datasets. R@i denotes Recall at i. All methods have been trained on ALIGN, JFT, LTIP for the same number of training steps.}
\label{tab:app_retrieval}
\end{table*}

\subsection{Hyperparameters details}

We train all models using the AdamW~\citep{loshchilov2017decoupled} optimizer, a cosine learning rate schedule with linear warm-up of 2500 steps. For all methods, we sweep over learning rate and weight decay values in the following ranges: learning rate in $[7e-4, 9e-4, 1.1 e-4]$ and weight decay in $[0.1, 0.2, 0.3]$. We use a batch size of 16348 (except for GLoRIA for which we use 4096 batch size) and we pre-train the ViT-B models for 200k steps ($\approx$ 3.2 billion data points).

For the other SPARC hyperparameters, we set the global loss weight $\lambda_g = 0.5$ and we sweep the local loss weight in $\lambda_f \in [0.5, 1.0, 5.0, 10.0]$. Moreover, we use a learned temperature parameter $\tau$. 

For baseline specific hyperparameters, we follow the publicly available code (where available) and the original papers.
For MGCA \citep{mgca}, as described in the paper,  we set the weighing of the different losses  $\lambda_1 = 1$, $\lambda_2 = 1$, $\lambda_3 = 1$, the number of attention heads for computing the cross-modal embeddings to 1 with a 128 embedding dimension. For MGCA's crossmodal prototype alignment loss, we use 500 prototypes with $\epsilon=0.05$ and 3 iterations for the Sinkhorn-Knopp clustering algorithm. 

For FILIP, we implemented the token dropping procedure described in the paper and use 20\% token dropping in our experiments.

For PACL, we closely follow the original paper in terms of implementation up to one notable detail -- we include a learnable temperature parameter in the loss as we found this to significantly improve performance.

\subsection{Prompt ensembling for zero-shot classification} \label{apx.prompt_ensambling}

Following \cite{clip} and \cite{filip} we use prompt templates to augment the label for classification tasks. We use the prompt templates format from \cite{filip}:
\begin{equation}
    [\text{prefix}] \{ \text{class label} \}, [\text{suffix}]
\end{equation}
For the [\text{prefix}], we use the templates from \cite{clip}. On the other hand, for the [\text{suffix}], we use the templates from \cite{filip}, which shows that adding the reference word `it' at the end of the prompt, e.g. `I like it', further improves performance.

\subsection{Memory consumption and FLOPS for the different methods} \label{apx.flops}

We provide detailed numbers for the FLOPS (in TFLOPS) and of the Peak Memory (in MB) in Table \ref{tab:flops_memory}.

\begin{table*}[h]
\begin{center}
\begin{adjustbox}{max width=1.0\columnwidth}
\begin{tabular}{lCCCCCCCC}  %
\toprule
 &   \multicolumn{4}{c}{FLOPS (TFLOPS)} & \multicolumn{4}{c}{Peak memory (MB)} \\
\text{Objective}  & \text{B = 2048} & \text{B = 4096} & \text{B = 8192} & \text{B = 16384} & \text{B = 2048} & \text{B = 4096} & \text{B = 8192} & \text{B = 16384} \\
\midrule
\midrule
CLIP & 1.15 & 2.29 & 4.57 & 9.14 & 4394 & 4452 & 5889 & 8578 \\
PACL & 1.2 & 2.46 & 5.24 & 12.8 & 4682 & 6267 & 9786 & 14785 	 \\
GLoRIA & 3.34 & 13.21 & - & - & 8013 & 13840 & - & -  \\ %
MGCA & 1.16 &  2.31 & 4.62 & 9.23 & 4412 & 4462 & 5936 & 8681 \\
FILIP & 1.37 & 3.17 & 8.09 & 27.25 & 4394 & 5230 & 8657 & 15463  \\
SPARC (ours) & 1.15 & 2.3 & 4.6 & 9.19 & 4408 & 4450 & 5914 & 8620  \\

\bottomrule
\end{tabular}
\end{adjustbox}
\end{center}
\caption{TFLOPS and peak memory usage for one update step of each method for different batch sizes. }
\label{tab:flops_memory}
\end{table*}

\subsection{Semantic segmentation}

For zero-shot semantic segmentation, we pass the 
patch embeddings through the extra dense layer and the adapter to compute the cosine similarity with the text embeddings for the ground-truth classes. Similarly to \citep{pacl} we compute the mean Intersection over Union (mIoU) only for the foreground classes.

\subsection{SPARC backbones in vision language models}

We train the Perceiver Resampler part of Flamingo \citep{flamingo} on the ALIGN \citep{align}, LTIP (Long Text \& Image Pairs)~\citep{flamingo} and VTP (Video \& Text Pairs) \citep{flamingo} datasets. VTP consists of 27 million short
videos paired with text descriptions, where each video if ~22s on average. We use the AdamW optimizer, a cosine learning rate schedule with peak learning rate of $1e-4$, linear warmup with 5000 warm-up steps and 250k training steps in total.

\subsection{SPARC vs CLIP Faithfulness Examples}

\begin{figure*}[h!]
    \centering
    \hspace*{-50pt}\includegraphics[width=1.2\columnwidth]{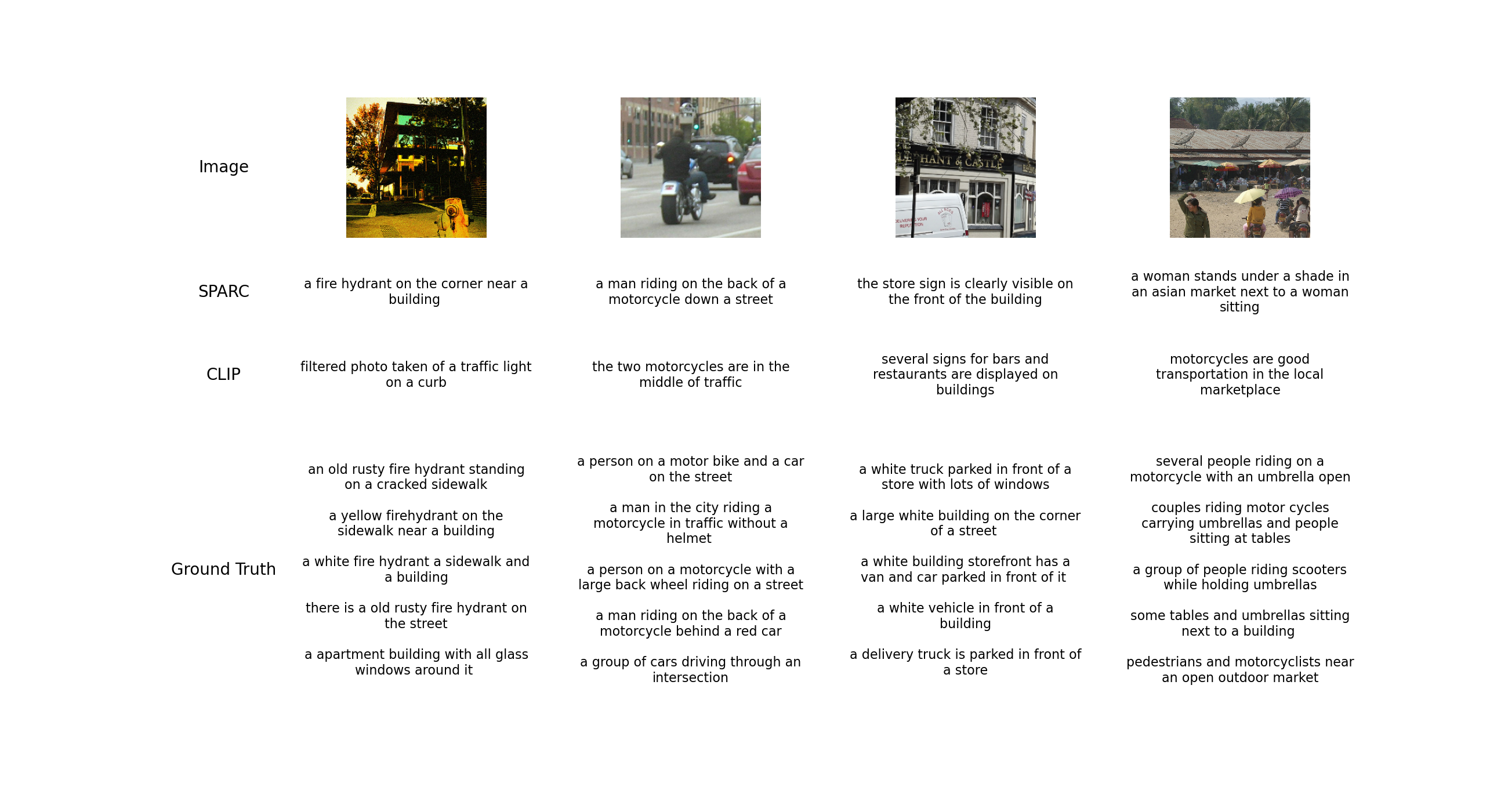}
    \vspace{-15mm}
    \caption{SPARC vs CLIP vs Ground Truth for examples where SPARC has higher all-token $\mathcal{K}$-Precision ($\mathcal{K}$-P)}
    \label{sup:examples-faithfulness-tokens}
\end{figure*}

\begin{figure*}[h!]
    \centering
    \hspace*{-50pt}\includegraphics[width=1.2\columnwidth]{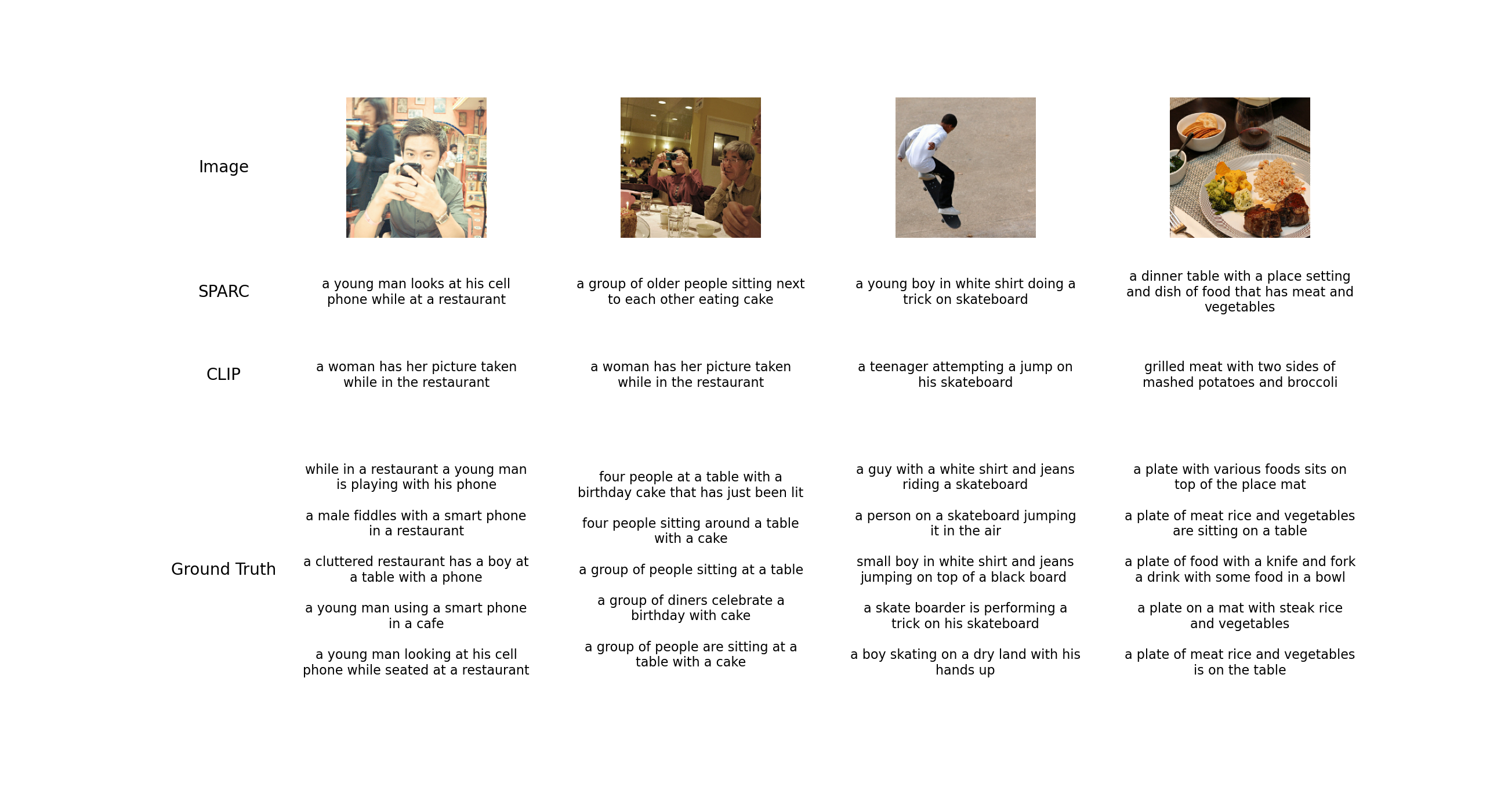}
    \vspace{-15mm}
    \caption{SPARC vs CLIP vs Ground Truth for examples where SPARC has higher $\mathcal{K}$-Precision restricted to nouns and adjectives ($\mathcal{K}$-P\textsubscript{na})}
    \label{sup:examples-faithfulness-nounish}
\end{figure*}

To further understand the ability of SPARC and CLIP models to faithfully describe the elements in the image, we provide several qualitative examples. Thus, for MSCOCO, we chose examples where the top-1 retrieved caption for both SPARC and CLIP is not part of the ground truth captions, but where where SPARC has higher all-token $\mathcal{K}$-Precision (Figure \ref{sup:examples-faithfulness-tokens}) and higher $\mathcal{K}$-Precision restricted to nouns and adjectives (\ref{sup:examples-faithfulness-nounish}). From these figure, we notice that captions retrieved using the CLIP representations describe objects that not present in the image (e.g. ``several signs for bars'' when there are none present) or get the number of objects wrong (e.g. "two motorcycles" when there is only one motorcycle). Alternatively, captions retrieved using the SPARC representations are more faithful to the image, but also provide more descriptive details (e.g. "young boy in white shirt", "dinner table with a place setting").

\newpage

\end{document}